\documentclass{article}

     \PassOptionsToPackage{numbers, sort&compress}{natbib}

\usepackage[main, final]{neurips_2025}

\usepackage[utf8]{inputenc} 
\usepackage[T1]{fontenc}    
\usepackage[colorlinks=true, citecolor=mydarkblue]{hyperref}       
\usepackage{url}            
\usepackage{booktabs}       
\usepackage{amsfonts}       
\usepackage{nicefrac}       
\usepackage{microtype}      
\usepackage[table,xcdraw,dvipsnames]{xcolor}        

\input{preamble}

\usepackage{amsmath,amsfonts,bm}

\def\eqref#1{equation~\ref{#1}}

\def\1{\bm{1}}

\DeclareMathAlphabet{\mathsfit}{\encodingdefault}{\sfdefault}{m}{sl}
\SetMathAlphabet{\mathsfit}{bold}{\encodingdefault}{\sfdefault}{bx}{n}

\usepackage{url}
\usepackage{amssymb}
\usepackage{times}
\usepackage{helvet}
\usepackage{courier}
\usepackage{graphicx}
\usepackage{natbib}
\usepackage{caption}
\usepackage{algorithm}
\usepackage{algorithmic}

\usepackage[utf8]{inputenc}
\usepackage[T1]{fontenc}
\usepackage{url}
\usepackage{booktabs}
\usepackage{amsfonts}
\usepackage{amsmath}
\usepackage{amsthm}
\usepackage{nicefrac}
\usepackage{microtype}
\usepackage{xspace}
\usepackage{wrapfig}
\usepackage{multirow}
\usepackage{makecell}
\usepackage{array}
\usepackage{enumitem}
\usepackage{comment}
\usepackage{pifont}
\usepackage{wrapfig}
\usepackage{subcaption}
\usepackage[font=small]{caption}
\usepackage[normalem]{ulem}
\useunder{\uline}{\ul}{}

\newcommand\barenote[1]{%
  \begingroup
  \renewcommand\thefootnote{}\footnote{#1}%
  \addtocounter{footnote}{-1}%
  \endgroup
}

\title{OVS Meets Continual Learning: Towards Sustainable Open-Vocabulary Segmentation}

\author{
\textbf{Dongjun Hwang}$^{1}$ \quad \textbf{Yejin Kim}$^{1}$ \quad \textbf{Minyoung Lee}$^{1}$ \quad \textbf{Seong Joon Oh}$^{2,3}$ \quad \textbf{Junsuk Choe}$^{1\dag}$ \vspace{.5em} \\
{\normalsize
$^1$Sogang University \xspace\xspace\xspace
$^2$University of Tübingen \xspace\xspace\xspace
$^3$Tübingen AI Center \xspace\xspace\xspace
}
}

\begin{document}

\maketitle

\begin{abstract}
Open-Vocabulary Segmentation (OVS) aims to segment classes that are not present in the training dataset. However, most existing studies assume that the training data is fixed in advance, overlooking more practical scenarios where new datasets are continuously collected over time. To address this, we first analyze how existing OVS models perform under such conditions. In this context, we explore several approaches such as retraining, fine-tuning, and continual learning but find that each of them has clear limitations. To address these issues, we propose ConOVS, a novel continual learning method based on a Mixture-of-Experts framework. ConOVS dynamically combines expert decoders based on the probability that an input sample belongs to the distribution of each incremental dataset. Through extensive experiments, we show that ConOVS consistently outperforms existing methods across pre-training, incremental, and zero-shot test datasets, effectively expanding the recognition capabilities of OVS models when data is collected sequentially. Code is available at: {\url{https://github.com/dongjunhwang/ConOVS}}
\end{abstract}
\barenote{\hspace{-2em}\scriptsize \dag \xspace Correspondence to \texttt{jschoe@sogang.ac.kr}.}

\section{Introduction}
\label{sec:introduction}

In fields such as robotics~\citep{gu2024conceptgraphs} and autonomous driving~\citep{wong2020identifying, li2022coda}, there is a growing demand for models that can segment novel objects not included in the training dataset. However, conventional closed-set segmentation models, which are restricted to recognizing only the classes seen during training, fall short in meeting this demand. To address this limitation, Open-Vocabulary Segmentation (OVS) has emerged, aiming to enable segmentation of unseen classes that are not included in the training dataset. OVS continues to be an active area of research, particularly through methods that leverage foundation models such as CLIP~\citep{maskclip, fcclip}.

Most previous studies~\citep{zhu2024survey, ODISE, fcclip, zhang2023simple} on OVS assume a scenario in which the model is trained once using a pre-training dataset. However, in practice, trainable datasets often arrive sequentially as new data are collected over time. Considering this setting, we first discuss how existing OVS models perform under such conditions. To facilitate a clearer discussion, we measure the relative performance of OVS models on seen and unseen classes using a \textit{reference baseline}. We adopt OneFormer~\citep{oneformer} for this role. It represents the state-of-the-art in closed-set segmentation and shares the same ConvNeXt backbone~\citep{convnext} as the OVS model~\citep{fcclip}, enabling a fair comparison in model capacity.

\begin{figure}[t]
    \centering
    \scriptsize
    \setlength{\columnsep}{10pt}
    \begin{subfigure}[b]{0.47\linewidth}
        \centering
        \includegraphics[width=\linewidth]{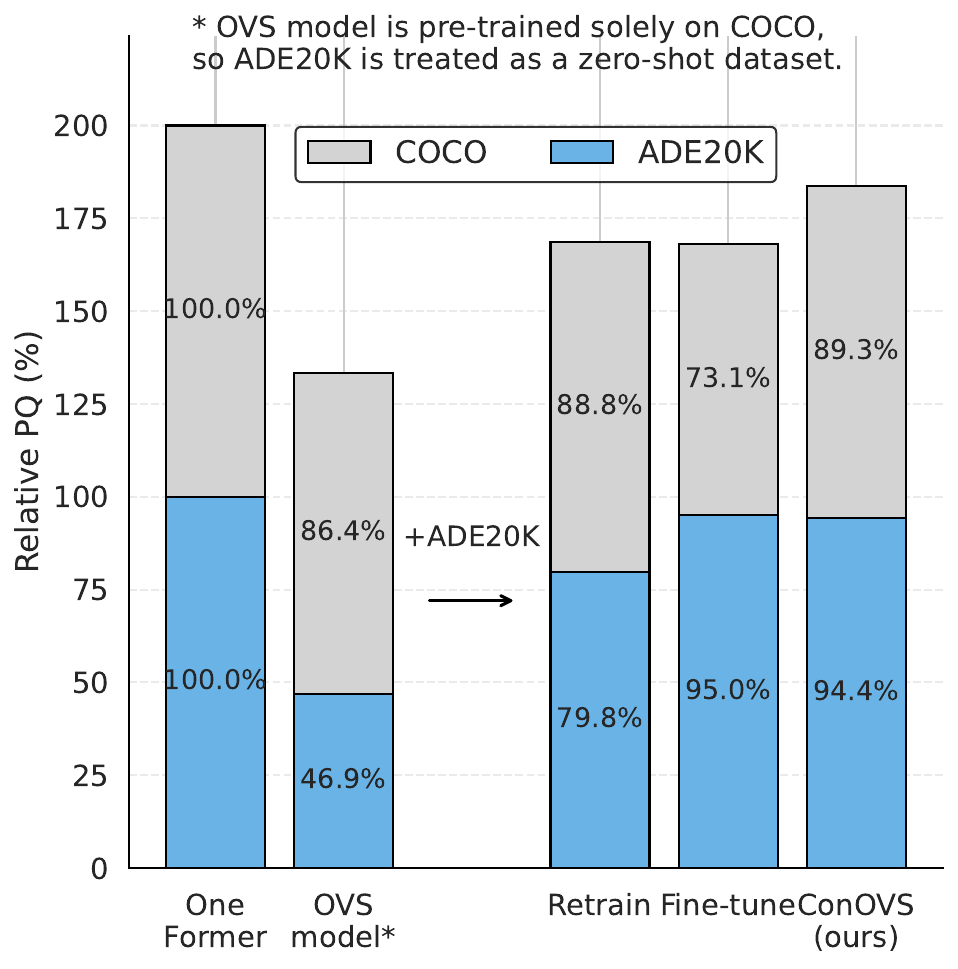}
        \caption{ }
        \label{fig:teaser_oneformer}
    \end{subfigure}
    \begin{subfigure}[b]{0.52\linewidth}
        \centering
        \includegraphics[width=\linewidth]{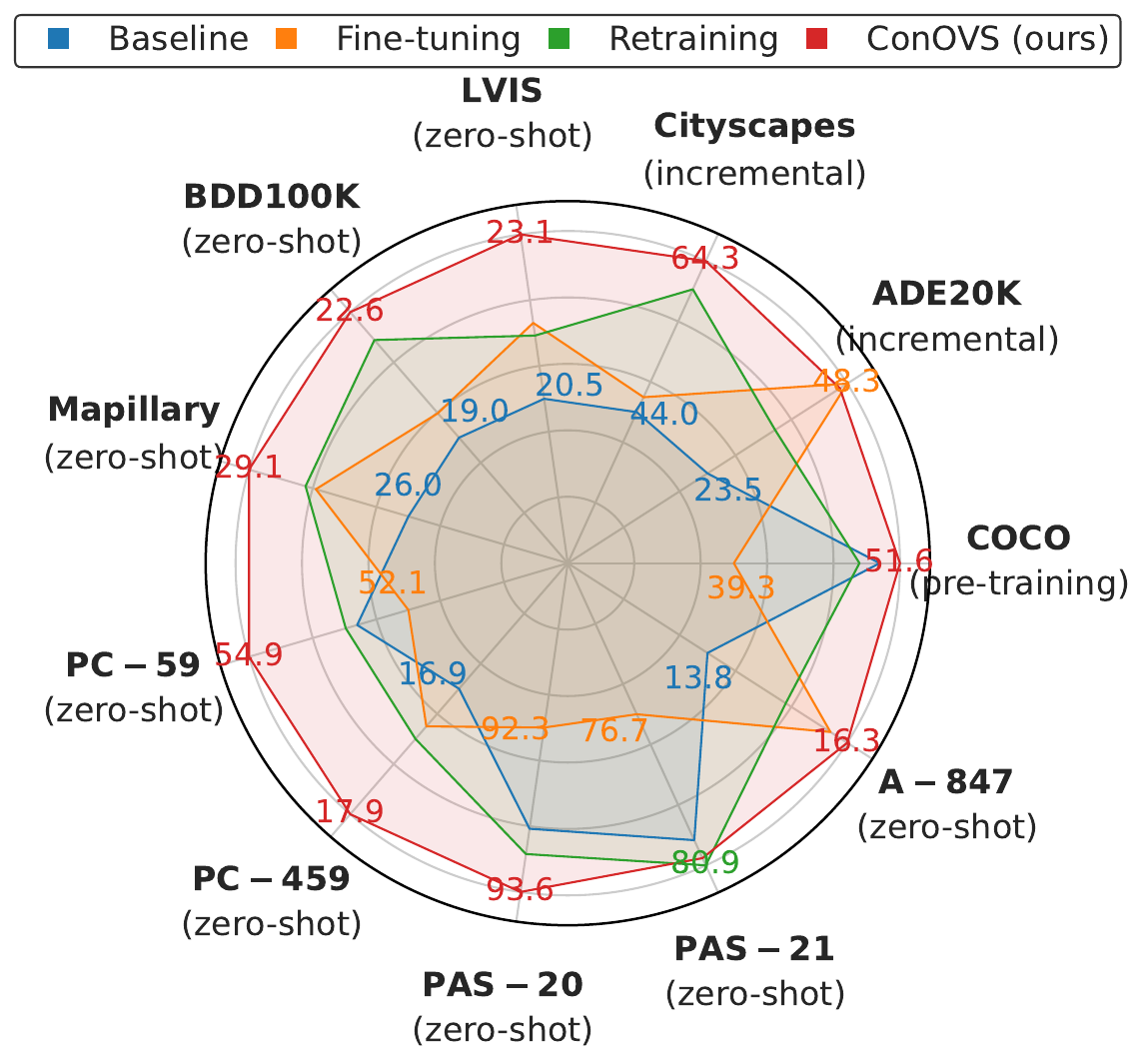}
        \caption{ }
        \label{fig:teaser_circle}
    \end{subfigure}
     \vspace{-.5em}
    \caption{\small (a) Comparison of the performance of the OVS model (fc-clip~\citep{fcclip}), Retraining, Fine-tuning, and ConOVS against the closed-set segmentation model OneFormer. (b) Performance of the Baseline (fc-clip~\citep{fcclip}), Fine-tuning, Retraining, and ConOVS on the pre-training, incremental, and zero-shot test datasets. \pq is used.}
    \label{fig:teaser}
     \vspace{-1.5em}
\end{figure}

The most straightforward approach is to use the existing OVS model as is.  In our experiments (Figure~\ref{fig:teaser_oneformer}), the existing OVS model achieves 86.4\% of OneFormer's performance on the pre-training dataset COCO~\citep{coco}. In contrast, its performance drops to 46.9\% on a new dataset ADE20K~\cite{ade20k}, which contains unseen classes as well. These results indicate that the OVS model fails to perform well on datasets it has not encountered during training.

To determine whether this performance gap is due to the inherent difficulty of the unseen classes or simply because the model has not been trained on them, we retrain the OVS model using both the pre-training dataset and the new dataset. As shown in Figure~\ref{fig:teaser_oneformer}, the model’s performance on ADE20K improves significantly from 46.9\% to 79.8\% relative to OneFormer. This result confirms that the low performance on the new dataset is primarily due to the lack of exposure during training. It also suggests that this limitation of the OVS model can be effectively mitigated by training on newly collected data. 

However, retraining the model from scratch demands substantial computational resources. In particular, this approach becomes impractical when the pre-training data is no longer accessible or computational resources are limited. To address these limitations, we consider an alternative approach: transfer learning. Specifically, we fine-tune the pre-trained OVS model on a new dataset. However, as shown in Figure~\ref{fig:teaser_oneformer} and~\ref{fig:teaser_circle}, this approach also has a limitation. It leads to performance degradation not only on the pre-training dataset but also on zero-shot tasks. This issue appears to stem from a well-known drawback of fine-tuning, namely, \textit{catastrophic forgetting}. Therefore, we consider continual learning (CL) methods, which are designed to address catastrophic forgetting. However, most CL approaches are developed under the assumption that the number of classes is finite, making them unsuitable for open-vocabulary tasks where the number of classes can be potentially infinite~\citep{CLinOVC, AnytimeCL}. 

As a result, in scenarios where new datasets are continuously collected—as assumed in this paper—it is still unclear how to effectively utilize the incoming data, and finding a viable solution in OVS remains a non-trivial challenge. To address this, we propose ConOVS, a Mixture-of-Experts (MoE) based continual learning method that incrementally trains an existing OVS model on new datasets. Our method begins by fine-tuning the pre-trained OVS model to build a distinct expert for each new dataset. During inference, we estimate the probability that a given input sample is close to the distribution of each training dataset, based on their statistical representations. The model then computes an interpolation factor from these probabilities and dynamically combines the experts by interpolating their weights. This allows our method to produce an optimal model for predicting each input sample.

To simulate the scenario assumed in this paper, we sequentially introduce incremental datasets to an existing OVS model and evaluate the resulting models on three validation sets: the pre-training dataset, the incremental dataset, and zero-shot datasets. As shown in Figure~\ref{fig:teaser_circle}, our method not only significantly improves performance on the incremental dataset compared to standard retraining and fine-tuning, but also consistently enhances performance on both the pre-training and zero-shot datasets. Furthermore, compared to existing continual learning methods, our approach achieves superior performance across all three evaluation settings.

\section{Related Works}

\subsection{Open-Vocabulary Segmentation}

Recent open-vocabulary segmentation (OVS) research has focused on leveraging models capable of open-vocabulary classification, such as CLIP~\citep{CLIP}, to recognize classes that are not included in the training dataset. For example, fc-clip~\citep{fcclip} identifies unseen classes by combining class embeddings from the model’s decoder with those from CLIP. Moreover, a recent study~\cite{semla} has explored an approach that retrieves LoRA modules trained on different datasets according to the input and utilizes them in conjunction with CLIP. Other methods further enhance the recognition of unseen classes by either applying visual grounding techniques like GradCAM~\citep{gradcam} to CLIP~\citep{maskclip, sun2024clip, luo2023segclip} or distilling knowledge from both CLIP and the segmentation foundation model Segment Anything Model (SAM)~\citep{OV-SAM, samclip}. Meanwhile, there are also OVS approaches that do not rely on CLIP. For instance, methods such as X-Decoder~\citep{XDecoder, SEEM, zhang2023simple} train both the encoder and decoder from scratch using segmentation datasets along with large-scale image–text pair datasets.

Most existing OVS studies are based on a scenario in which the model is trained only once. However, this setting inherently limits performance on unseen classes (see Section~\ref{sec:introduction}). To overcome this limitation, we analyze strategies for training OVS models in a scenario where new datasets are introduced sequentially.

\subsection{Continual Learning}

Acquiring additional knowledge in an already trained model is not straightforward. When a model is further trained on new data, it often tends to forget previously learned information while learning the new content~\citep{mccloskey1989catastrophic}. This phenomenon is widely known as \textit{catastrophic forgetting}. To address this issue, the field of continual learning (CL) has emerged. CL explores methods that enable models to learn from new data while retaining prior knowledge.

CL techniques are typically categorized into three types. First, replay-based methods store a subset of previously seen data and retrain the model using it to preserve prior knowledge~\citep{ERcontinual, cha2021ssul}. Second, regularization-based methods introduce penalty terms in the loss function to constrain parameter updates, preventing significant deviations during training on a new dataset~\citep{lwf, ewc, MiB}. Third, parameter-isolation-based methods mitigate interference by freezing previously learned parameters and allocating separate parameters for learning new data~\citep{l2p, eclipse}. Several approaches extend this idea into a Mixture-of-Experts (MoE) framework, where additional parameter sets are treated as distinct experts, and a gating module selects the appropriate expert based on the input~\citep{le2024mixture, sprompt}.

However, existing CL methods are designed under the assumption that the number of classes is finite, which limits their applicability in open-vocabulary settings~\citep{CLinOVC, AnytimeCL, zheng2023preventing}. Therefore, there is a need for novel approaches that enable continual learning in Open-Vocabulary Segmentation (OVS) scenarios, where new data are introduced incrementally. To address this, we propose a novel MoE-based continual learning technique that effectively expands the capacity of OVS models.

\section{Motivation}
\label{sec:motivation}

In this section, we expand on the discussion from Section~\ref{sec:introduction} and explore in greater detail how newly collected datasets can be leveraged to improve the performance of OVS models.

The most straightforward approach is to retrain the model from scratch using a joint dataset that combines the original and newly collected data. In practice, this strategy effectively preserves performance on seen classes while substantially improving performance on unseen classes. However, it suffers from two major limitations: (1) it incurs significant computational costs, as the model must be retrained from scratch every time new data are added; and (2) retraining becomes entirely infeasible if access to the original dataset has expired.

\begin{figure}[t]
    \centering
    \scriptsize
    \setlength{\columnsep}{10pt}
    
    \begin{subfigure}[t]{0.52\linewidth}
        \centering
        \includegraphics[width=\linewidth]{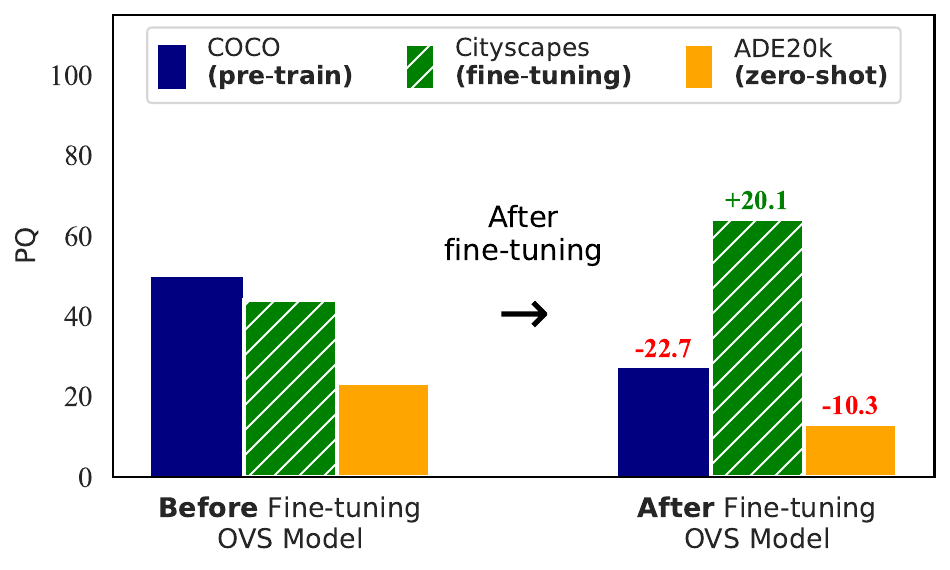}
        \caption{ }
        \label{fig:finetuning}
    \end{subfigure}%
    \hfill
    \begin{subfigure}[t]{0.45\linewidth}
        \centering
        \includegraphics[width=\linewidth]{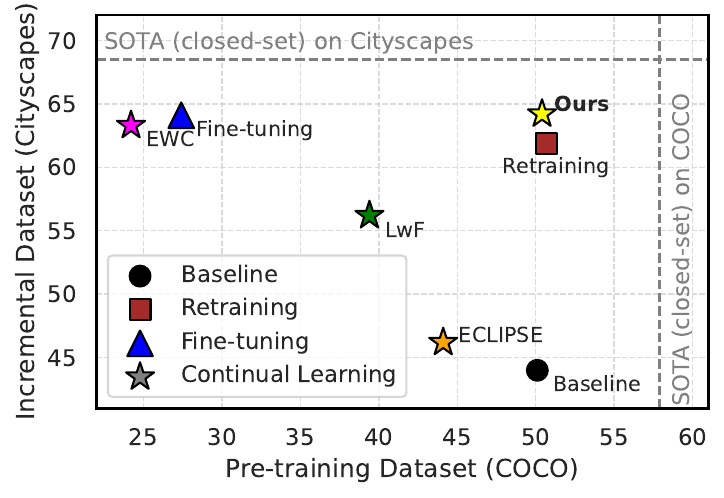}
        \caption{ }
        \label{fig:cl_comparison}
    \end{subfigure}

    \caption{\small (a) Performance degradation on the pre-training and zero-shot datasets after fine-tuning. fc-clip is used. (b) Comparison of the performance of OneFormer~\cite{oneformer}, the baseline (fc-clip~\cite{fcclip}), retraining, fine-tuning, three existing continual learning methods~\cite{ewc, lwf, eclipse}, and ConOVS on the pre-training and incremental datasets. All methods use the same iterations. \pq is used.}
    \vspace{-1.em}
    \label{fig:quantitative_combined}
\end{figure}

Due to these limitations, fine-tuning the model using only the newly collected dataset may appear to be a practical alternative. However, this approach compromises the model's original performance. As shown in Figure~\ref{fig:finetuning}, fine-tuning the OVS model results in a significant drop in performance not only on the pre-training dataset but also on the zero-shot test dataset. Qualitative examples provided in the Appendix~\ref{app:qualitative} further illustrate this phenomenon. This degradation is likely caused by a well-known issue in fine-tuning, known as catastrophic forgetting~\citep{lwf, ewc}.

Another potential direction is to apply continual learning (CL) methods to OVS models. However, most existing CL methods are built on the assumption of a finite set of classes, making them difficult to directly apply to open-vocabulary tasks~\citep{CLinOVC, AnytimeCL, zheng2023preventing}. For instance, \cite{MiB, eclipse} apply CL to segmentation tasks by treating all unseen classes as background, which fundamentally conflicts with the goal of OVS models that aim to recognize potentially unlimited categories.

Even when existing CL methods are adapted for OVS (see Appendix~\ref{app:implementation_detail} for implementation details), our experimental results show that their effectiveness is limited. As shown in Figure~\ref{fig:cl_comparison}, OVS models trained with adapted CL methods perform significantly worse than the closed-set segmentation model OneFormer on both the pre-training and incremental datasets. We believe this arises because existing CL methods assume a closed-set segmentation with a finite label space, whereas OVS involves a potentially infinite label space, which these methods do not account for.

To address these issues, we propose \textbf{ConOVS}, a new continual learning method that sequentially improves the performance of OVS models. Specifically, ConOVS (1) reduces training cost by using only newly collected data, unlike retraining; (2) avoids catastrophic forgetting, unlike fine-tuning; and (3) effectively improves performance on the incremental and zero-shot test dataset, unlike existing CL methods.

\section{Background}
\label{sec:background}

\myparagraph{Open-Vocabulary Segmentation (OVS)} aims to predict segmentation mask–class pairs from an input image $\bm{x}_{\text{img}}$ and a text description $\bm{x}_{\text{text}}$, which may include both seen (trained) and unseen classes. OVS models typically consist of three components: an image encoder, a text encoder, and a decoder, denoted as $f = \{f_{\text{img}}, f_{\text{text}}, f_{\text{dec}}\}$. The image encoder $f_{\text{img}}$ produces an image embedding $\bm{z}_{\text{img}}$, and the text encoder $f_{\text{text}}$ produces a text embedding $\bm{z}_{\text{text}}$. These are fed into the decoder $f_{\text{dec}}$, which, given $N$ learnable object queries, outputs $N$ pairs of predicted masks and class embeddings, $\{(\bm{m}_i, \bm{c}_i)\}_{i=1}^{N}$. Each $\bm{m}_i$ is a predicted mask, and $\bm{c}_i$ is its associated class embedding. Final class labels are assigned by matching each $\bm{c}_i$ to the most similar text embedding.

\myparagraph{Continual Learning Setup.}
We consider a continual learning scenario in which datasets containing new classes arrive sequentially, and the set of seen classes gradually expands over time. The model $f$ is first trained on a pre-training dataset $\mathcal{D}_{\text{pre}}$, and then incrementally updated using a sequence of datasets $\mathcal{D}_{\text{inc},1}, \mathcal{D}_{\text{inc},2}, \cdots$. At each time step $t \in \{1, 2, \cdots, n\}$, the model is trained only on $\mathcal{D}_{\text{inc},t}$, without access to $\mathcal{D}_{\text{pre}}, \mathcal{D}_{\text{inc},1}, \cdots, \mathcal{D}_{\text{inc},t-1}$. The class set $\mathcal{C}_t$ from each incremental dataset is added to the previously seen class set, resulting in $\mathcal{C}_{\text{seen}} = \bigcup_{s=1}^{t} \mathcal{C}_s \cup \mathcal{C}_{\text{pre}}$. The model is evaluated on the test sets of all datasets up to time $t$ to assess both its ability to learn new classes and retain prior knowledge. To additionally evaluate generalization, we use a zero-shot test set $\mathcal{D}_{\text{zero}}$ containing unseen classes $\mathcal{C}_{\text{unseen}} \subset \mathcal{C}_{\text{total}} \setminus \mathcal{C}_{\text{seen}}$ that never appeared during training.

\section{The Proposed Method: ConOVS}
\label{sec:methodology}

In this section, we propose \textbf{ConOVS}, a novel MoE-based continual learning method designed to train OVS models in scenarios where new datasets are sequentially collected. For clarity, we describe the proposed method in two parts: \textit{Training Phase} and \textit{Inference Phase}.

\subsection{Training Phase}

During training, we derive \textit{expert models} and \textit{multivariate normal} (MVN) \textit{distributions} for each dataset. Specifically, we first train an OVS model from scratch using the pre-training dataset. Then, we fine-tune only the decoder on each incremental dataset to obtain an expert model specific to that dataset. For each dataset, we also compute the mean and covariance matrix of the image and text embeddings, which define the MVN distributions. These are represented as $\bm{\Phi}^i_{\text{img}} = (\bm{\mu}^i_{\text{img}}, \bm{\Sigma}^i_{\text{img}})$ and $\bm{\Phi}^i_{\text{text}} = (\bm{\mu}^i_{\text{text}}, \bm{\Sigma}^i_{\text{text}})$ for each dataset $i \in \{0, \cdots, n\}$. Here, $i = 0$ corresponds to the pre-training dataset, while $i \in \{1, \cdots, n\}$ refers to the incremental datasets.

\subsection{Inference Phase}

We perform inference by dynamically combining expert models based on the MVN distributions derived during training. Specifically, we first compute task vectors $\bm{v}_i$ for each expert model, defined as the arithmetic difference between the decoder weights of the $i$-th incremental expert $\bm{\theta}_{\text{dec,inc}}^i$ and the pre-trained decoder weights $\bm{\theta}_{\text{dec,pr}}$. Given an input sample, we feed the image $\bm{x}_{\text{img}}$ and class descriptions $\bm{x}_{\text{text}}$ into the image and text encoders, respectively, to obtain the corresponding embeddings $\bm{z}_{\text{img}}$ and $\bm{z}_{\text{text}}$. We then evaluate the likelihoods of these embeddings under the MVN distributions for all datasets, and collect them into the vectors $\bm{l}_{\text{img}}, \bm{l}_{\text{text}} \in \mathbb{R}^{n+1}$. 

\begin{figure}[t]
    \centering
    \begin{minipage}[h]{0.58\textwidth}
        \centering
        \includegraphics[width=\textwidth]{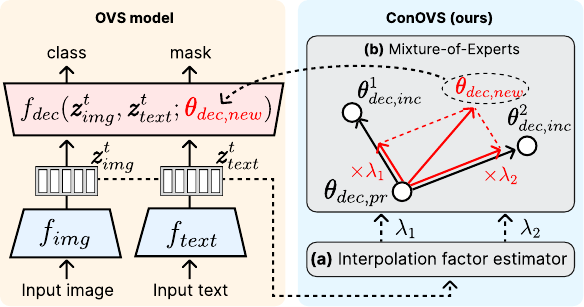}
        \caption{\small Overview of the inference process of our proposed method.}
        \label{fig:overview_inf}
    \end{minipage}
    \hfill
    \begin{minipage}[h]{0.4\textwidth}
        \vspace{-1.4em}
        \centering
        \begin{algorithm}[H] 
            \small
            \caption{\small Interpolation factor estimator}
            \begin{algorithmic}[1]
                \REQUIRE Input $(\bm{x}_{\text{img}}, \bm{x}_{\text{text}})$, encoders $f_{\text{img}}, f_{\text{text}}$, decoder $f_{\text{dec}}$; MVN parameters $\{\bm{\Phi}_{\text{img}}^i, \bm{\Phi}_{\text{text}}^i\}_{i=0}^{n}$; PDF $p(\cdot|\bm{\Phi})$
                \ENSURE Interpolation factor $\bm{\lambda}$
                \STATE Extract embeddings: $\bm{z}_{\text{img}} \gets f_{\text{img}}(\bm{x}_{\text{img}})$, $\bm{z}_{\text{text}} \gets f_{\text{text}}(\bm{x}_{\text{text}})$
                \STATE Estimate likelihoods: $\bm{l}_{\text{img}} \gets \{p(\bm{z}_{\text{img}} \mid \bm{\Phi}_{\text{img}}^i)\}$, $\bm{l}_{\text{text}} \gets \{p(\bm{z}_{\text{text}} \mid \bm{\Phi}_{\text{text}}^i)\}$
                \STATE Compute: $\bm{p}_{\text{img}} \gets \text{softmax}(\bm{l}_{\text{img}})$, $\bm{p}_{\text{text}} \gets \text{softmax}(\bm{l}_{\text{text}})$
                \STATE Combine: $\bm{\lambda} \gets \max(\bm{p}_{\text{img}}, \bm{p}_{\text{text}})$
                \RETURN $\bm{\lambda}$
            \end{algorithmic}
            \label{alg:interpolation_factor_estimator}
        \end{algorithm}
    \end{minipage}
\end{figure}

After that, we apply the softmax operation to the log-likelihood vector to normalize the proximity scores of each domain into the $[0,1]$ range. This decision is motivated by a prior study~\cite{ilharco2022editing}, which reported that merging performance degrades when the interpolation factor exceeds 1. Finally, we compute the element-wise maximum of the two probability vectors to obtain the final interpolation factor vector $\bm{\lambda} \in \mathbb{R}^{n+1}$. The detailed procedure is provided in Algorithm~\ref{alg:interpolation_factor_estimator}, and ablation studies on the choice of softmax and element-wise maximum are presented in Appendix~\ref{app:additional_ablation}.

The final decoder weights $\bm{\theta}_{\text{dec,new}}$ are computed as:
\begin{equation}
\bm{\theta}_{\text{dec,new}} = \bm{\theta}_{\text{dec,pr}} + \sum_{i=1}^{n} \lambda_i \bm{v}_i.
\end{equation}%
That is, the decoder is dynamically constructed by linearly combining task vectors $\bm{v}_i$ with interpolation weights $\lambda_i$, relative to the pre-trained decoder (see Figure~\ref{fig:overview_inf}\redcolor{b}). Note that while $\lambda_0$ is not directly used in this computation, it is included in the softmax operation and thus indirectly affects the other $\bm{\lambda}$ elements. As a result, when the input is close to the pre-training distribution, $\lambda_0$ approaches 1, pushing the remaining $\lambda_i$ values toward 0.

The effectiveness and justification of this design are empirically validated in Section~\ref{sec:experiments}.

\section{Experiments}
\label{sec:experiments}

\myparagraph{Learning Sequences.}
This study assumes a scenario where trainable datasets arrive sequentially and evaluates OVS models that are incrementally trained on them. In the main paper, we examine three learning sequences. In Scenario 1 \textbf{(S1)}, the model is pre-trained on COCO~\citep{coco}, incrementally trained on Cityscapes~\cite{cityscapes}, and evaluated on ADE20K~\cite{ade20k} as the zero-shot test set. In Scenario 2 \textbf{(S2)}, the model is again pre-trained on COCO but incrementally trained on ADE20K, with Cityscapes used for zero-shot evaluation. In Scenario 3 \textbf{(S3)}, the model is pre-trained on COCO and incrementally trained on both Cityscapes and ADE20K. For zero-shot evaluation, we use a diverse collection of datasets: LVIS~\citep{lvis}, BDD100K~\citep{bdd100k}, Mapillary Vistas~\citep{mapillary}, PC-59, PC-459~\citep{pascal_context}, PAS-20, PAS-21~\cite{pascalvoc}, and A-847~\cite{ade20k}. We further validate our method on a larger number of incremental datasets in Scenario 4 (\textbf{S4}), with the results provided in Appendix~\ref{app:more_incremental}. Evaluation is conducted on the test sets of the pre-training and incremental datasets, as well as the designated zero-shot test sets.

\myparagraph{Implementation Details.} We apply our method to two OVS models: fc-clip with ConvNeXt-L~\citep{convnext} and X-Decoder with Focal-L~\citep{focal}. During the pre-training phase, fc-clip trains only the decoder, while X-Decoder trains both the encoder and decoder. In the fine-tuning phase, both models train only the decoder. The temperature $T$ in the softmax is set to 0.01, and log-likelihood is used to compute probabilities from the MVN distributions. All experiments are run on two NVIDIA A5000 GPUs.

\myparagraph{Evaluation Metrics.} We evaluate panoptic, instance, and semantic segmentation using PQ, mAP, and mIoU, respectively. Due to space constraints, we report only PQ in the main paper, with the others in the Appendix~\ref{app:all_quantitative}. Some zero-shot test datasets support only specific segmentation tasks; for example, LVIS supports only instance segmentation. In such cases, we evaluate performance only on the supported task.

\subsection{Main Results}
\label{sec:main_results}

\begin{table*}[t]
	\centering
	\caption{\small Comparison of performance across Baselines (fc-clip, X-Decoder), Retraining, Fine-tuning, four existing continual learning methods, and ConOVS when the incremental dataset is (a) Cityscapes or (b) ADE20K. \pq is used.}
	\begin{minipage}{0.48\textwidth}
		\centering
            \subcaption{\small Cityscapes\label{tab:cityscapes}}
		\scriptsize
\setlength{\tabcolsep}{0.4em}
\renewcommand{\arraystretch}{1.1}
\begin{tabular}{@{}lcc|c|c@{}}
\textbf{Method} & \textbf{\begin{tabular}[c]{@{}c@{}}CL\end{tabular}} & \textbf{\begin{tabular}[c]{@{}c@{}}COCO \\ (\pretraining)\end{tabular}} & \textbf{\begin{tabular}[c]{@{}c@{}}Cityscapes \\ (incremental)\end{tabular}} & \textbf{\begin{tabular}[c]{@{}c@{}}ADE20K \\ (\unseen)\end{tabular}} \\
\addlinespace[2pt]
\Xhline{3\arrayrulewidth}
\addlinespace[3pt]
fc-clip & \xmark & 50.1 & 44.0 & 23.5 \\ \midrule
Fine-tuning & \xmark & \tableminus{\tableminus{-22.7}} & \tableplus{+20.1} & \tableminus{-10.3} \\
\fix{Retraining} & \xmark & \textbf{\tableplus{+0.6}} & \tableplus{+17.9} & \tableplus{+1.7} \\
ER & \cmark & \tableminus{-1.6} & \tableplus{+19.0} & \tableplus{+0.3} \\
LwF & \cmark & \tableminus{-10.7} & \tableplus{+12.2} & \tableminus{-0.8} \\
EWC & \cmark & \tableminus{\tableminus{-25.9}} & \tableplus{+19.3} & \tableminus{-9.8} \\
ECLIPSE & \cmark & \tableminus{-6.0} & \tableplus{+2.2} & \tableplus{+0.9} \\
\rowcolor[HTML]{EFEFEF}
\textbf{ConOVS (ours)} & \cmark & \tableplus{+0.3} & \textbf{\tableplus{+20.2}} & \textbf{\tableplus{+2.5}} \\ \midrule
X-Decoder & \xmark & 56.7 & 36.3 & 16.7 \\ \midrule
Fine-tuning & \xmark & \tableminus{-50.4} & \textbf{\tableplus{+26.6}} & \tableminus{\tableminus{-12.9}} \\
\rowcolor[HTML]{EFEFEF}
\textbf{ConOVS (ours)} & \cmark & \textbf{\tableminus{-0.4}} & \textbf{\tableplus{+26.6}} & \textbf{\tableplus{+0.1}}
\end{tabular}

	\end{minipage}
	\hfill
	\begin{minipage}{0.48\textwidth}
		\centering
            \subcaption{\small ADE20K\label{tab:ade20k}}
		\scriptsize
\setlength{\tabcolsep}{0.4em}
\renewcommand{\arraystretch}{1.1}
\begin{tabular}{@{}lcc|c|c@{}}
\textbf{Method} & \textbf{\begin{tabular}[c]{@{}c@{}}CL\end{tabular}} & \textbf{\begin{tabular}[c]{@{}c@{}}COCO \\ (\pretraining)\end{tabular}} & \textbf{\begin{tabular}[c]{@{}c@{}}ADE20K \\ (incremental)\end{tabular}} & \textbf{\begin{tabular}[c]{@{}c@{}}Cityscapes \\ (\unseen)\end{tabular}} \\
\addlinespace[2pt]
\Xhline{3\arrayrulewidth}
\addlinespace[3pt]
fc-clip & \xmark & 50.1 & 23.5 & 44.0 \\ \midrule
Fine-tuning & \xmark & \tableminus{-7.7} & \textbf{\tableplus{+24.1}} & \tableminus{-3.0} \\
\fix{Retraining} & \xmark & \tableplus{+1.4} & \tableplus{+16.5} & \tableminus{-1.2} \\
ER & \cmark & \tableplus{+0.4} & \tableplus{+21.5} & \tableminus{-3.5} \\
LwF & \cmark & \tableminus{-3.8} & \tableplus{+13.7} & \tableminus{-1.0} \\
EWC & \cmark & \tableminus{\tableminus{-11.1}} & \tableplus{+20.7} & \tableminus{-2.6} \\
ECLIPSE & \cmark & \tableminus{-0.5} & \tableplus{+0.2} & \tableminus{-5.9} \\
\rowcolor[HTML]{EFEFEF}
\textbf{ConOVS (ours)} & \cmark & \textbf{\tableplus{+1.7}} & \tableplus{+23.8} & \textbf{\tableplus{+0.9}} \\ \midrule
X-Decoder & \xmark & 56.7 & 16.7 & 36.3 \\ \midrule
Fine-tuning & \xmark & \tableminus{\tableminus{-37.3}} & \tableplus{+28.2} & \tableminus{-3.7} \\
\rowcolor[HTML]{EFEFEF}
\textbf{ConOVS (ours)} & \cmark & \textbf{\tableminus{-1.5}} & \textbf{\tableplus{+29.2}} & \textbf{\tableplus{+1.4}}
\end{tabular}

	\end{minipage}
	\label{tab:single}
\end{table*}

In this section, we compare the performance of the proposed ConOVS and other approaches under the three scenarios. We first analyze the results for scenarios S1 and S2, followed by scenario S3. We then provide a more in-depth analysis of our method, including an investigation into the behavior of the interpolation factors. All methods were trained with the same number of iterations to ensure a fair comparison, and detailed information on the training cost of each method is provided in Appendix~\ref{app:training_resources}.

In scenarios \textbf{S1} and \textbf{S2}, where only a single incremental dataset is used for training, our method consistently outperforms existing approaches across all datasets, whether the incremental dataset is ADE20K or Cityscapes (see Table~\ref{tab:single}). In particular, compared to retraining, our method almost maintains or even improves performance on the pre-training dataset, despite not using it during additional training (e.g., Retraining: +1.4 vs. Ours: +1.7 in S2). It also achieves superior performance on the incremental dataset itself (e.g., Retraining: +16.5 vs. Ours: +23.8 in S2). Moreover, compared to fine-tuning and conventional continual learning, our method improves performance on the incremental dataset without compromising performance on the pre-training dataset. This improvement is attributed to the dynamic interpolation of expert models in our method, which helps mitigate catastrophic forgetting.

Our method also achieves the best performance on the zero-shot test dataset. For instance, in scenario S2, performance on the Cityscapes improves by +0.9, whereas all other methods show performance drops. This result indicates that our method enhances recognition of a wider range of classes while preserving previously learned knowledge.

\begin{wraptable}{r}{0.60\textwidth}
		\centering
		\scriptsize
		\setlength{\tabcolsep}{.2em}
		\renewcommand{\arraystretch}{1.2}
            \vspace{-1.em}
		\caption{Performance comparison in scenario S3. The best performance for each dataset is underlined. “City$\rightarrow$ADE” means fine-tuning on Cityscapes first, then ADE20K. \pq is used.}
		\scriptsize
		\begin{tabular}{@{}lcc|c|c@{}}
 \textbf{Method} & \textbf{\begin{tabular}[c]{@{}c@{}} Learning \\ Sequence\end{tabular}} & \textbf{\begin{tabular}[c]{@{}c@{}}COCO \\ (\pretraining)\end{tabular}} & \textbf{\begin{tabular}[c]{@{}c@{}}ADE20K \\ (incremental)\end{tabular}} & \textbf{\begin{tabular}[c]{@{}c@{}}Cityscapes \\ (incremental)\end{tabular}} \\ \midrule \midrule
fc-clip & - & 50.1 & 23.5 & 44.0 \\ \midrule
Fine-tuning & ADE → City & 20.8 & 15.4 & {\ul 65.2} \\
Fine-tuning & City → ADE & 39.3 & {\ul 48.3} & 46.0 \\
Retraining & COCO, City, ADE & 48.6 & 35.5 & 60.5 \\
\rowcolor[HTML]{EFEFEF}
\textbf{ConOVS (ours)} & City, ADE & {\ul \textbf{51.6}} & \textbf{47.0} & \textbf{64.3}
\end{tabular}
		\label{tab:multiple}
\end{wraptable}

In scenario \textbf{S3}, our method consistently achieves superior performance compared to both fine-tuning and retraining. Specifically, as shown in Table~\ref{tab:multiple}, fine-tuning performs well only on the most recently trained dataset, whereas our method consistently achieves strong results on all three datasets. By contrast, retraining shows lower performance than our method, likely due to its need for more iterations to converge. In comparison, our method yields better results with the same number of training iterations, demonstrating greater training efficiency. Note that the analysis related to the number of training iterations in retraining is provided in Appendix~\ref{app:iter_retraining}.

\begin{table*}[h]
\centering
\scriptsize
\setlength{\tabcolsep}{0.8em}
\renewcommand{\arraystretch}{1.2}
\caption{\small Performance comparison on 8 unseen datasets in scenario S3. The best performance for each dataset is underlined. \pq is used.}
\begin{tabular}{@{}lccccccccc@{}}
\textbf{Method} & \textbf{\begin{tabular}[c]{@{}c@{}}Learning \\ Sequence\end{tabular}} & \textbf{\begin{tabular}[c]{@{}c@{}}LVIS\\ (\map)\end{tabular}} & \textbf{\begin{tabular}[c]{@{}c@{}}BDD100K\\ (\pq)\end{tabular}} & \textbf{\begin{tabular}[c]{@{}c@{}}Mapillary\\ (\miou)\end{tabular}} & \textbf{\begin{tabular}[c]{@{}c@{}}PC-59\\ (\miou)\end{tabular}} & \textbf{\begin{tabular}[c]{@{}c@{}}PC-459\\ (\miou)\end{tabular}} & \textbf{\begin{tabular}[c]{@{}c@{}}PAS-20\\ (\miou)\end{tabular}} & \textbf{\begin{tabular}[c]{@{}c@{}}PAS-21\\ (\miou)\end{tabular}} & \textbf{\begin{tabular}[c]{@{}c@{}}A-847\\ (\miou)\end{tabular}} \\ \midrule \midrule
fc-clip & - & 20.5 & 19.0 & 26.0 & 53.0 & 16.9 & 93.1 & 80.2 & 13.8 \\ \midrule
Fine-tuning & City → ADE & 21.7 & 19.7 & 27.8 & 52.1 & 17.2 & 92.3 & 76.7 & 16.0 \\
Fine-tuning & ADE → City & 10.4 & 21.3 & 24.2 & 45.9 & 13.5 & 87.4 & 70.7 & 11.5 \\
\fix{Retraining} & COCO, City, ADE & 21.5 & 21.8 & 28.0 & 53.2 & 17.3 & 93.3 & {\ul 80.9} & 15.2 \\
\rowcolor[HTML]{EFEFEF}
\textbf{ConOVS (ours)} & City, ADE & {\ul \textbf{23.1}} & {\ul \textbf{22.6}} & {\ul \textbf{29.1}} & {\ul \textbf{54.9}} & {\ul \textbf{17.9}} & {\ul \textbf{93.6}} & \textbf{80.7} & {\ul \textbf{16.3}}
\end{tabular}
\label{tab:multiple_unseen}
\end{table*}

In addition, our method also consistently outperforms other approaches in various zero-shot evaluations. As shown in Table~\ref{tab:multiple_unseen}, it achieves superior performance across all eight zero-shot test datasets. This result suggests that the dynamic interpolation of expert models in our method facilitates recognition of a broader range of unseen classes.

\begin{table}[h]
	\centering
	\scriptsize
	\setlength{\tabcolsep}{1.em}
	\renewcommand{\arraystretch}{1.1}
        \caption{\small Comparison of performance on seen and unseen classes in the zero-shot test dataset ADE20K. \miou is used. (b) Comparison of PQ, SQ, and RQ between fc-clip and ConOVS in the zero-shot test dataset ADE20K.}
	\begin{subtable}[t]{0.45\linewidth}
		\centering
		\caption{ }
		\begin{tabular}{lcc}
\toprule
\textbf{Method} & \textbf{Seen Classes} & \textbf{Unseen Classes} \\ \midrule
fc-clip & 35.0 (+0.0) & 28.6 (+0.0) \\
\rowcolor[HTML]{EFEFEF} \textbf{ConOVS (ours)} & \textbf{37.9 (\tableplus{+2.9})} & \textbf{30.9 (\tableplus{+2.3})} \\
\bottomrule
\end{tabular}
            \label{tab:truly_unseen}
	\end{subtable}%
	\hfill
	\begin{subtable}[t]{0.53\linewidth}
		\centering
		\caption{ }
		\begin{tabular}{lccc}
\toprule
Method & PQ & SQ & RQ \\
\midrule
fc-clip        & 23.5 (+0.0) & 61.7 (+0.0) & 28.3 (+0.0) \\
\rowcolor[HTML]{EFEFEF}
ConOVS (ours) & \textbf{25.9 (\tableplus{+2.4})} & \textbf{73.1 (\tableplus{+11.4})} & \textbf{31.2 (\tableplus{+2.9})} \\
\bottomrule
\end{tabular}
		\label{tab:reason_unseen}
	\end{subtable}
	\label{tab:unseen_combined}
\end{table}

\myparagraph{Evaluation of the Truly Unseen Classes.} Some classes in the zero-shot test datasets may overlap with those in the training data. For instance, ADE20K shares 38 of its 150 classes with COCO. To more accurately assess zero-shot performance, we separately evaluate the model on truly unseen classes that do not appear in the training data. Therefore, we split ADE20K into seen and unseen subsets and measure performance on each in scenario S1.

As shown in Table~\ref{tab:truly_unseen}, our method improves performance by a similar margin on both seen and unseen classes (seen: +2.9, unseen: +2.3). This suggests that the performance gain is not solely from improved recognition of seen classes, but also reflects better generalization to unseen classes.

\myparagraph{Analysis of Improvements in Unseen Classes.} To better understand the source of performance improvements in unseen classes, we analyzed results on the zero-shot dataset ADE20K by comparing the PQ, SQ, and RQ scores of the baseline and our proposed method. As shown in Table~\ref{tab:reason_unseen}, incorporating ConOVS into the baseline model improves both PQ and RQ. The most notable gain, however, is observed in SQ, which evaluates the quality of the predicted segmentation masks. These results indicate that the improvements in unseen classes are primarily driven by enhanced segmentation quality rather than improved mask classification.

\begin{figure}
	\centering
	\includegraphics[width=\textwidth]{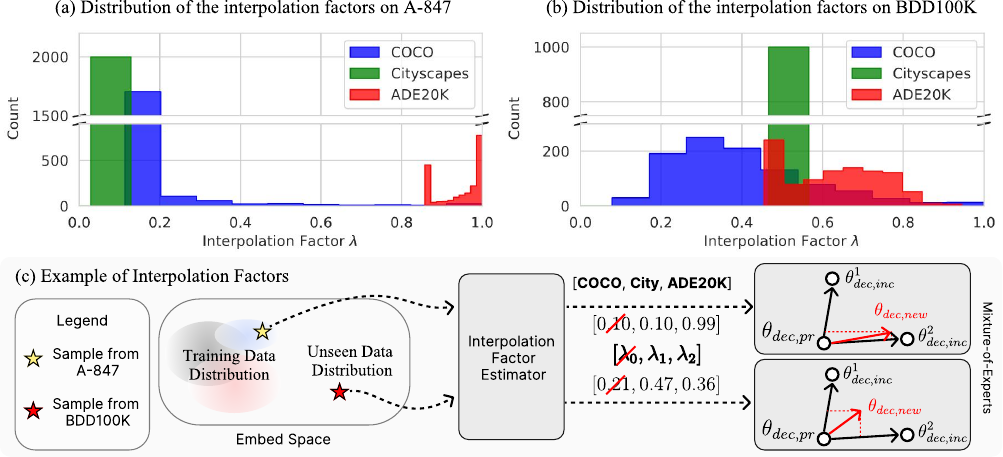}
	\caption{\small
		Interpolation factor behavior across different input sample distributions.
	}
	\label{fig:factor_behavior}
\end{figure}

\myparagraph{Understanding the Behavior of the Interpolation Factor.} We analyze how the proposed method adapts to different input sample distributions. To this end, we examine the distribution of interpolation factors $\bm{\lambda}$ estimated by the interpolation factor estimator across two zero-shot test datasets. One is A-847, which shares a similar distribution with the incremental training dataset ADE20K, and the other is BDD100K, which differs significantly from all training datasets.

As shown in Figure~\ref{fig:factor_behavior}\redcolor{a}, the interpolation factors for A-847 tend to be close to 0 or 1. In particular, the expert trained on ADE20K receives a $\lambda$ value close to 1, while other experts receive values close to 0. This shows that when input samples are similar to a previously trained distribution, our method selectively activates the corresponding expert to maximize performance (see Figure~\ref{fig:factor_behavior}\redcolor{c} top-right).

In contrast, as illustrated in Figure~\ref{fig:factor_behavior}\redcolor{b}, the interpolation factors for BDD100K are more evenly distributed between 0 and 1. This suggests that the input samples do not clearly belong to any of the known training distributions. In such cases, our method disperses the $\lambda$ values to avoid over-reliance on a single expert. Instead, it combines the weights of multiple experts based on the probability that the input sample belongs to each distribution. This allows the model to leverage knowledge from various datasets and produce more accurate predictions even for samples from unfamiliar domains (see Figure~\ref{fig:factor_behavior}\redcolor{c} bottom-right).

\subsection{Ablation Study}

In this section, we conduct ablation studies to analyze the contribution of each component in the proposed method. All experiments are conducted in scenario S1.

\begin{table}[t]
	\centering
	\scriptsize
	\setlength{\tabcolsep}{.5em}
	\renewcommand{\arraystretch}{1.2}
        \caption{(a) Comparison of the interpolation factor estimator when using both image and text distributions versus using only one of them. \pq is used. (b) Performance comparison when the MVN distribution is replaced with K-means clustering or KDE. fc-clip and \pq are used.}
	\begin{subtable}[t]{0.45\linewidth}
		\centering
		\caption{ }
		\fixall{
\begin{tabular}{@{}lc|c|c@{}}
\toprule
 \textbf{\begin{tabular}[c]{@{}c@{}} Distribution\end{tabular}} & \textbf{\begin{tabular}[c]{@{}c@{}}COCO \\ (\pretraining)\end{tabular}} & \textbf{\begin{tabular}[c]{@{}c@{}}Cityscapes \\ (incremental)\end{tabular}} & \textbf{\begin{tabular}[c]{@{}c@{}}ADE20K\\(zero-shot)\end{tabular}} \\ \midrule
image only & 51.5 & 43.4 & 25.8 \\
text only & \textbf{51.9} & 60.7 & 25.9 \\
\rowcolor[HTML]{EFEFEF}
image + text & 51.6 & \textbf{64.3} & 26.0 \\ \bottomrule
\end{tabular}
}

		\label{tab:image_text}
	\end{subtable}%
	\hfill
	\begin{subtable}[t]{0.53\linewidth}
		\centering
		\caption{ }
		\fixall{
\begin{tabular}{@{}lc|c|c@{}}
\toprule
 \textbf{\begin{tabular}[c]{@{}c@{}} Methods\end{tabular}} & \textbf{\begin{tabular}[c]{@{}c@{}}COCO \\ (\pretraining)\end{tabular}} & \textbf{\begin{tabular}[c]{@{}c@{}}Cityscapes \\ (incremental)\end{tabular}} & \textbf{\begin{tabular}[c]{@{}c@{}}ADE20K\\(zero-shot)\end{tabular}} \\ \midrule
k-means clustering & 42.4 & 64.1 & 26.1 \\
kernel density estimation & 48.1 & 57.4 & 26.1 \\
\rowcolor[HTML]{EFEFEF}
\begin{tabular}[l]{@{}c@{}}MVN distribution\end{tabular} & \textbf{50.4} & \textbf{64.3} & 26.0 \\ \bottomrule
\end{tabular}
}

		\label{tab:prototype}
	\end{subtable}
	\label{tab:ablation_combined}
\end{table}

\myparagraph{Ablation Study of Image and Text Distribution.} Our method computes the interpolation factor of an input sample using the MVN distributions of image and text embeddings for each training dataset. To analyze how the interpolation factors are affected by the distribution design, we compare three configurations: image only, text only, and combined image-text.

As shown in Table~\ref{tab:image_text}, using both image and text distributions yields the best performance on the incremental dataset. This suggests that combining both modalities enables more accurate estimation of the input sample’s proximity to training distributions, leading to better expert selection.

\myparagraph{Evaluating Alternative Approaches against the MVN Distribution.} We evaluate and compare two alternative techniques to the MVN distribution used in our method for estimating interpolation factors. Specifically, we replace the MVN distribution with K-means clustering or Kernel Density Estimation (KDE), and analyze the resulting performance changes. Detailed descriptions of the K-means and KDE are provided in the Appendix~\ref{app:comparison_description_mvn}.

As shown in Table~\ref{tab:prototype}, both K-means and KDE yield lower performance on the pre-training and incremental dataset. These results suggest that the MVN distribution enables more accurate estimation of interpolation factors for in-distribution data. We attribute this to its relatively simple structure and low dimensionality, which make it less sensitive to outliers than K-means or KDE.

\myparagraph{Replacing Softmax with Argmax.} The proposed method uses the softmax function to compute interpolation factors for each dataset. We compare the performance on eight zero-shot datasets when replacing the softmax function with the argmax operation. Table~\ref{tab:softmax_argmax} presents the evaluation results. The experimental results show that softmax consistently outperforms argmax across all zero-shot datasets (e.g., on LVIS, argmax: $21.3$, softmax: $23.1$).

\begin{table}[ht]
\scriptsize
\setlength{\tabcolsep}{0.8em}
\renewcommand{\arraystretch}{1.2}
\caption{\small Performance comparison between the argmax and softmax operations in the interpolation factor estimator. We use fc-clip with our method and fine-tune it on both Cityscapes and ADE20K. \pq is used.}
\begin{tabular}{@{}lccccccccc@{}}
\textbf{Decision Rule} & \textbf{\begin{tabular}[c]{@{}c@{}}Incremental \\ Dataset \end{tabular}} & \textbf{\begin{tabular}[c]{@{}c@{}}LVIS\\ (\map)\end{tabular}} & \textbf{\begin{tabular}[c]{@{}c@{}}BDD100K\\ (\pq)\end{tabular}} & \textbf{\begin{tabular}[c]{@{}c@{}}Mapillary\\ (\miou)\end{tabular}} & \textbf{\begin{tabular}[c]{@{}c@{}}PC-59\\ (\miou)\end{tabular}} & \textbf{\begin{tabular}[c]{@{}c@{}}PC-459\\ (\miou)\end{tabular}} & \textbf{\begin{tabular}[c]{@{}c@{}}PAS-20\\ (\miou)\end{tabular}} & \textbf{\begin{tabular}[c]{@{}c@{}}PAS-21\\ (\miou)\end{tabular}} & \textbf{\begin{tabular}[c]{@{}c@{}}A-847\\ (\miou)\end{tabular}} \\ \midrule \midrule
Argmax & Cityscapes, ADE20k & 21.3 & 18.3 & 26.9 & 53.1 & 17.0 & 93.2 & 80.2 & \textbf{16.3} \\
\rowcolor[HTML]{EFEFEF}
Softmax & Cityscapes, ADE20k & \textbf{23.1} & \textbf{22.6} & \textbf{29.1} & \textbf{54.9} & \textbf{17.9} & \textbf{93.6} & \textbf{80.7} & \textbf{16.3} \\
\end{tabular}
\label{tab:softmax_argmax}
\end{table}

Specifically, on datasets such as LVIS and BDD100K, softmax demonstrates clearly superior performance. However, for PAS-20, PAS-21, and A-847, the performance difference between softmax and argmax is minimal. This occurs because, when the input sample is close to the distribution of the pre-training or incremental dataset, the interpolation factor obtained from softmax tends to be close to 0 or 1. As a result, softmax behaves similarly to argmax.

\begin{wraptable}{r}{0.5\textwidth}
    \centering
    \scriptsize
    \setlength{\tabcolsep}{.5em}
    \renewcommand{\arraystretch}{1.2}
    \vspace{-1.em}
    \caption{Effect of softmax temperature $T$ on performance across datasets. \miou is used.}
    \begin{tabular}{ccccc} \hline \textbf{$T$} & \textbf{\makecell{COCO\\(pre-training)}} & \textbf{\makecell{ADE20K\\(incremental)}} & \textbf{\makecell{Cityscapes\\(zero-shot)}} & \textbf{Total} \\ \hline \hline
 0.0001 & 50.7 & 35.4 & 43.8 & 129.9 \\
 0.001 & 51.2 & 42.2 & \textbf{43.9} & 137.3 \\ 
0.01 & \textbf{51.8} & 47.3 & 43.7 & \textbf{142.8} \\ 
0.1 & 51.3 & \textbf{47.5} & 43.2 & 142.0 \\ 
1.0 & 51.2 & 47.4 & 43.2 & 141.8 \\ \hline 
\end{tabular}
    \label{tab:temperature_analysis}
\end{wraptable}

\myparagraph{Hyperparameter Sensitivity Analysis.} Our method uses a softmax operation to compute the interpolation factor, and we analyze the effect of the softmax temperature hyperparameter $T$. The temperature $T$ directly influences the distribution of the interpolation factor: a low $T$ smooths the factor values, while a high $T$ pushes them toward extreme values of 0 or 1. Table~\ref{tab:temperature_analysis} summarizes how this behavior affects performance.

When $T$ is small, the interpolation factor $\lambda$ becomes overly smoothed, which prevents the expert models for each dataset from being utilized. This leads to performance degradation on the incremental dataset. In contrast, when $T$ is large, $\lambda$ converges to values close to 0 or 1, resulting in the selective use of a single expert model. This degrades performance on the zero-shot dataset. These findings suggest that appropriately integrating multiple models is essential for effective generalization to zero-shot datasets, and that extreme interpolation factors hinder this process.

\myparagraph{Decoder Interpolation.} Unlike our method, which fine-tunes the entire decoder for each dataset, existing MoE-based continual learning methods~\citep{le2024mixture, sprompt} primarily adopt Visual Prompt Tuning (VPT), where only a small subset of parameters is trained for each incremental dataset. This approach differs from ours in two key aspects: expert models consist of only partial decoder parameters, and a single expert is selected at inference time instead of performing interpolation. To assess the effectiveness of our full decoder fine-tuning strategy, we replace it with the VPT-based approach and compare their performance.

\begin{wraptable}{r}{0.5\textwidth}
		\centering
		\scriptsize
		\setlength{\tabcolsep}{.5em}
		\renewcommand{\arraystretch}{1.2}
            \vspace{-1.em}
		\caption{\small Performance comparison when the decoder interpolation in our method is replaced with a visual prompt tuning-based approach. fc-clip and \pq are used.}
		\scriptsize
		\fixall{
\begin{tabular}{@{}lc|c|c@{}}
 \textbf{\begin{tabular}[c]{@{}c@{}}Method\end{tabular}} & \textbf{\begin{tabular}[c]{@{}c@{}}COCO \\ (\pretraining)\end{tabular}} & \textbf{\begin{tabular}[c]{@{}c@{}}Cityscapes \\ (incremental)\end{tabular}} & \textbf{\begin{tabular}[c]{@{}c@{}}ADE20K\\(zero-shot)\end{tabular}} \\ \midrule \midrule
Prompt Tuning & 43.3 & 48.9 & 24.4 \\
\rowcolor[HTML]{EFEFEF}
Decoder Interpolation & \textbf{50.4} & \textbf{64.3} & \textbf{26.0}
\end{tabular}
}

		\label{tab:prompt}
	\end{wraptable}

Specifically, we implement the prompt tuning method based on \cite{sprompt} as follows: (1) for each incremental dataset, we train only the decoder’s object queries and positional embeddings and store them in a prompt pool; (2) during inference, we compute interpolation factors for each dataset using the same procedure as our method; (3) we identify the dataset with the highest interpolation factor; and (4) retrieve the corresponding object queries and positional embeddings from the prompt pool and apply them to the decoder for prediction.

As shown in Table~\ref{tab:prompt} and the experimental results, the prompt tuning variant consistently underperforms our method across pre-training, incremental, and zero-shot test datasets. This suggests that full decoder fine-tuning enables more effective adaptation to new datasets compared to VPT, which is constrained by its limited number of trainable parameters. Moreover, interpolating multiple experts provides greater flexibility and representational power than selecting a single expert, further supporting the advantage of our approach.

\section{Limitation}

Our method generates a unique decoder weight for each input sample, which can limit its applicability when the inference batch size exceeds one—a common constraint in other MoE-based continual learning approaches~\citep{codaprompt, sprompt}. However, since only the decoder varies per input and the encoder is shared across samples, the encoder can process inputs in batches. The resulting embeddings are then decoded individually using their corresponding weights. This design reduces the batch size limitation by supporting batched encoder processing and per-sample decoding.

\section{Conclusion}

This paper identifies the performance limitations of existing Open-Vocabulary Segmentation (OVS) methods on unseen data, an aspect that has been largely overlooked in prior work. To address this issue, we introduce a new learning scenario in which newly collected datasets are incrementally used to further train the OVS model. Under this setting, we show that conventional approaches---such as retraining, fine-tuning, and continual learning---are either impractical or difficult to apply effectively.

To overcome these challenges, we propose \textbf{ConOVS}, a novel MoE-based continual learning method for OVS. In ConOVS, predictions are made by dynamically combining the decoders of expert models based on the probability that the input sample belongs to the distribution of each training dataset. We validate the effectiveness of our method through extensive evaluations across various sequential learning scenarios and compare it against existing approaches. Experimental results show that ConOVS consistently achieves superior performance on pre-training, incremental, and zero-shot test datasets, demonstrating its ability to effectively expand the recognition capability of OVS models.

\myparagraph{Broader Impacts.} The proposed method can be applied to real-world applications such as robotics, where new objects continuously appear in the environment. However, if the pre-training dataset is biased, the model may continue to produce skewed predictions even after additional training, as it is explicitly designed to preserve previously learned knowledge. It is therefore important to be aware of this characteristic of the proposed technique, as a lack of such awareness may lead to unexpected model behavior.

\section*{Acknowledgement}
We would like to thank Yeji Park, Beomyun Kwon, and Joonkyung Kim for the insightful discussions and valuable feedback during the development of this work. This work was partly supported by the Institute of Information \& Communications Technology Planning \& Evaluation (IITP) grant funded by the Korea government(MSIT) (No. RS-2025-25441313, Professional AI Talent Development Program for Multimodal AI Agents, Contribution: 50\%) and the National Research Foundation of Korea (NRF) grant funded by the Korea government (MSIT) (No. RS-2024-00350430, Mitigating Hallucinations for Trustworthy Large Vision-Language Model: Datasets, Evaluation, Learning, and Inference, Contribution: 50\%).

{
    \small
    \bibliographystyle{plainnat}
    \bibliography{ref}
}

\clearpage
\section*{NeurIPS Paper Checklist}

The checklist is designed to encourage best practices for responsible machine learning research, addressing issues of reproducibility, transparency, research ethics, and societal impact. Do not remove the checklist: {\bf The papers not including the checklist will be desk rejected.} The checklist should follow the references and follow the (optional) supplemental material.  The checklist does NOT count towards the page
limit. 

Please read the checklist guidelines carefully for information on how to answer these questions. For each question in the checklist:
\begin{itemize}
    \item You should answer \answerYes{}, \answerNo{}, or \answerNA{}.
    \item \answerNA{} means either that the question is Not Applicable for that particular paper or the relevant information is Not Available.
    \item Please provide a short (1–2 sentence) justification right after your answer (even for NA). 
\end{itemize}

{\bf The checklist answers are an integral part of your paper submission.} They are visible to the reviewers, area chairs, senior area chairs, and ethics reviewers. You will be asked to also include it (after eventual revisions) with the final version of your paper, and its final version will be published with the paper.

The reviewers of your paper will be asked to use the checklist as one of the factors in their evaluation. While "\answerYes{}" is generally preferable to "\answerNo{}", it is perfectly acceptable to answer "\answerNo{}" provided a proper justification is given (e.g., "error bars are not reported because it would be too computationally expensive" or "we were unable to find the license for the dataset we used"). In general, answering "\answerNo{}" or "\answerNA{}" is not grounds for rejection. While the questions are phrased in a binary way, we acknowledge that the true answer is often more nuanced, so please just use your best judgment and write a justification to elaborate. All supporting evidence can appear either in the main paper or the supplemental material, provided in appendix. If you answer \answerYes{} to a question, in the justification please point to the section(s) where related material for the question can be found.

IMPORTANT, please:
\begin{itemize}
    \item {\bf Delete this instruction block, but keep the section heading ``NeurIPS Paper Checklist"},
    \item  {\bf Keep the checklist subsection headings, questions/answers and guidelines below.}
    \item {\bf Do not modify the questions and only use the provided macros for your answers}.
\end{itemize}


\begin{enumerate}

\item {\bf Claims}
    \item[] Question: Do the main claims made in the abstract and introduction accurately reflect the paper's contributions and scope?
    \item[] Answer: \answerYes{} 
    \item[] Justification: The abstract and introduction provide a clear and accurate account of the paper’s main contributions and scope.

\item {\bf Limitations}
    \item[] Question: Does the paper discuss the limitations of the work performed by the authors?
    \item[] Answer: \answerYes{} 
    \item[] Justification: The limitations are clearly discussed in a separate Limitations section.

\item {\bf Theory assumptions and proofs}
    \item[] Question: For each theoretical result, does the paper provide the full set of assumptions and a complete (and correct) proof?
    \item[] Answer: \answerNA{} 
    \item[] Justification: Our paper does not include theoretical results.

    \item {\bf Experimental result reproducibility}
    \item[] Question: Does the paper fully disclose all the information needed to reproduce the main experimental results of the paper to the extent that it affects the main claims and/or conclusions of the paper (regardless of whether the code and data are provided or not)?
    \item[] Answer: \answerYes{} 
    \item[] Justification: We provide all necessary details, including hyperparameters and training setups, to ensure reproducibility of the main results.

\item {\bf Open access to data and code}
    \item[] Question: Does the paper provide open access to the data and code, with sufficient instructions to faithfully reproduce the main experimental results, as described in supplemental material?
    \item[] Answer: \answerYes{} 
    \item[] Justification:  We disclose all necessary details for reproducibility, including code, and training scripts in the supplementary materials.

\item {\bf Experimental setting/details}
    \item[] Question: Does the paper specify all the training and test details (e.g., data splits, hyperparameters, how they were chosen, type of optimizer, etc.) necessary to understand the results?
    \item[] Answer: \answerYes{} 
    \item[] Justification: We detail all training and evaluation settings, including data splits, backbones, and hyperparameters, along with the rationale behind their choices.

\item {\bf Experiment statistical significance}
    \item[] Question: Does the paper report error bars suitably and correctly defined or other appropriate information about the statistical significance of the experiments?
    \item[] Answer: \answerNo{} 
    \item[] Justification: We do not report error bars since running multiple trials for every experimental setup would require substantial computational resources.

\item {\bf Experiments compute resources}
    \item[] Question: For each experiment, does the paper provide sufficient information on the computer resources (type of compute workers, memory, time of execution) needed to reproduce the experiments?
    \item[] Answer: \answerYes{} 
    \item[] Justification: In the implementation details section, we provide the necessary information to reproduce our experiments. All experiments were conducted using two NVIDIA A5000 GPUs.
    
\item {\bf Code of ethics}
    \item[] Question: Does the research conducted in the paper conform, in every respect, with the NeurIPS Code of Ethics \url{https://neurips.cc/public/EthicsGuidelines}?
    \item[] Answer: \answerYes{} 
    \item[] Justification: We rigorously follow the NeurIPS Code of Ethics in all aspects of our research.

\item {\bf Broader impacts}
    \item[] Question: Does the paper discuss both potential positive societal impacts and negative societal impacts of the work performed?
    \item[] Answer: \answerYes{} 
    \item[] Justification: Both potential positive and negative societal impacts are discussed in the Broader Impact section. 
    
\item {\bf Safeguards}
    \item[] Question: Does the paper describe safeguards that have been put in place for responsible release of data or models that have a high risk for misuse (e.g., pretrained language models, image generators, or scraped datasets)?
    \item[] Answer: \answerNA{} 
    \item[] Justification: Our paper does not involve the release of any models or datasets with high risk of misuse.

\item {\bf Licenses for existing assets}
    \item[] Question: Are the creators or original owners of assets (e.g., code, data, models), used in the paper, properly credited and are the license and terms of use explicitly mentioned and properly respected?
    \item[] Answer: \answerYes{} 
    \item[] Justification: All external assets used in this work are publicly available and properly credited, with licenses and usage terms respected.

\item {\bf New assets}
    \item[] Question: Are new assets introduced in the paper well documented and is the documentation provided alongside the assets?
    \item[] Answer: \answerYes{} 
    \item[] Justification: We ensured that the model, code are well documented for clarity and reproducibility.

\item {\bf Crowdsourcing and research with human subjects}
    \item[] Question: For crowdsourcing experiments and research with human subjects, does the paper include the full text of instructions given to participants and screenshots, if applicable, as well as details about compensation (if any)? 
    \item[] Answer: \answerNA{} 
    \item[] Justification: Our work does not involve human subjects or crowdsourcing.

\item {\bf Institutional review board (IRB) approvals or equivalent for research with human subjects}
    \item[] Question: Does the paper describe potential risks incurred by study participants, whether such risks were disclosed to the subjects, and whether Institutional Review Board (IRB) approvals (or an equivalent approval/review based on the requirements of your country or institution) were obtained?
    \item[] Answer: \answerNA{} 
    \item[] Justification: Our paper does not involve crowdsourcing nor research with human subjects.

\item {\bf Declaration of LLM usage}
    \item[] Question: Does the paper describe the usage of LLMs if it is an important, original, or non-standard component of the core methods in this research? Note that if the LLM is used only for writing, editing, or formatting purposes and does not impact the core methodology, scientific rigorousness, or originality of the research, declaration is not required.
    \item[] Answer: \answerNA{} 
    \item[] Justification: Our work does not use LLMs in any important or non-standard way.

\end{enumerate}

\clearpage
\appendix
\section*{Technical Appendices and Supplementary Material}

\renewcommand\thefigure{\thesection\arabic{figure}} 
\setcounter{figure}{0}
\renewcommand{\thetable}{A\arabic{table}}
\setcounter{table}{0}

\section{Detailed Experimental Settings}
\label{app:experiment_setting}

\begin{table*}[ht]
\centering
\scriptsize
\setlength{\tabcolsep}{1.2em} 
\renewcommand{\arraystretch}{1.2}
\caption{\small Dataset configurations for pre-training, incremental, and zero-shot test datasets used in the four different learning sequences.
}
\begin{tabular}{@{}lccc@{}}
    \toprule
    \textbf{Type of Learning Sequence} & \textbf{Pre-training Dataset} & \textbf{Incremental Dataset} & \textbf{Zero-shot test Dataset} \\ \midrule \midrule
    (\textbf{S1}) Scenario 1 & COCO & Cityscapes & ADE20K \\
    (\textbf{S2}) Scenario 2 & COCO & ADE20K & Cityscapes \\
    (\textbf{S3}) Scenario 3 & COCO & Cityscapes, ADE20K & 
    \begin{tabular}[c]{@{}c@{}}LVIS, BDD100K, Mapillary Vistas, PC-59,\\ PC-459, PAS-20, PAS-21, A-847\end{tabular} \\
    (\textbf{S4}) Scenario 4 & COCO & 
    \begin{tabular}[c]{@{}c@{}}Cityscapes, ADE20K,\\ BDD100K, Mapillary Vistas\end{tabular} & 
    \begin{tabular}[c]{@{}c@{}}LVIS, PC-59, PC-459,\\ PAS-20, PAS-21, A-847\end{tabular} \\
    \bottomrule
\end{tabular}
\label{tab:dataset_config}
\end{table*}

This study assumes scenarios where trainable datasets are provided sequentially and evaluates the performance of OVS models that are incrementally trained on these datasets. Specifically, we first train an OVS model from scratch using the pre-training dataset. In the case of fc-clip, only the decoder is trained from scratch, whereas X-Decoder trains both the encoder and decoder. After pre-training, we incrementally train the model on the training sets of each incremental dataset in sequence. As shown in Table~\ref{tab:dataset_config}, we define four experimental scenarios (S1, S2, S3, S4) based on different learning sequences of the datasets. Finally, the model is evaluated using the evaluation sets of both the pre-training and incremental datasets, as well as zero-shot test datasets.

\subsection{Datasets}
\myparagraph{Pre-training Datasets.} In our experiments, we follow the pre-training dataset configurations proposed in each OVS model. Specifically, fc-clip uses only the COCO~\citep{coco} dataset for pre-training, while X-Decoder is trained on COCO for segmentation and additionally uses four image-text pair datasets. Since COCO is the only segmentation dataset used in the training of both models, we evaluate the segmentation performance on the pre-training dataset using the COCO evaluation set.

\myparagraph{Incremental Datasets.} We use Cityscapes~\cite{cityscapes} and ADE20K~\citep{ade20k} as incremental datasets. Cityscapes is specialized for urban driving scenes, whereas ADE20K covers a wide range of fine-grained things and stuff classes from both indoor and outdoor environments. In scenarios S1 and S2, where only one of these datasets is used for incremental learning, the other is used for zero-shot testing. For example, when Cityscapes is designated as the incremental dataset, ADE20K serves as the zero-shot test dataset.

\myparagraph{Zero-shot Test Datasets.} To evaluate the generalization performance of the OVS model on unseen classes not included during training, we use a total of eight datasets: LVIS~\citep{lvis}, BDD100K~\citep{bdd100k}, Mapillary Vistas~\citep{mapillary}, PC-59, PC-459, PAS-20, PAS-21, and A-847. Here, \textbf{PC} refers to Pascal Context~\citep{pascal_context}, \textbf{PAS} to Pascal VOC~\citep{pascalvoc}, and \textbf{A} to ADE20K. The number following each name indicates the number of classes included in the corresponding dataset.

\subsection{Implementation Details}
\label{app:implementation_detail}

We apply the proposed method to two OVS models: fc-clip with ConvNeXt-L~\citep{convnext} and X-Decoder with Focal-L~\citep{focal}. When fine-tuning on incremental datasets, we follow the original fine-tuning protocols of fc-clip and X-Decoder, in which the encoder is frozen and only the decoder is updated. The temperature parameter $T$ used in the softmax operation for estimating interpolation factors is set to 0.01. When computing probabilities from the MVN distributions, we use the log-likelihood. All experiments are conducted using two NVIDIA A5000 GPUs.

\subsection{Evaluation Metrics}
\label{sec:eval_metrics}

The two baselines used in this study, fc-clip and X-Decoder, are universal segmentation models capable of performing panoptic, instance, and semantic segmentation. Accordingly, we evaluate model performance across all three tasks using the following metrics: (1) Panoptic Segmentation is evaluated with Panoptic Quality (\pq), (2) Instance Segmentation with mean Average Precision (\map), and (3) Semantic Segmentation with mean Intersection over Union (\miou). All \pq, \miou, and \map values are reported in Table~\ref{tab:cityscapes_all},~\ref{tab:ade20k_all}, and~\ref{tab:multiple_all}, where these three metrics show consistent performance trends. In addition, some zero-shot test datasets are limited to specific segmentation types. For instance, LVIS supports only instance segmentation, while PC-459 provides only semantic segmentation annotations. In such cases, we evaluate performance using only the supported task for each dataset.

\section{Compared Methods}
\label{app:comparison_description}

\subsection{Overview of Adapted Continual Learning Methods for OVS}

Existing continual learning (CL) methods are known to be unsuitable for open-vocabulary tasks, as they are based on the assumption that only classes included in the training dataset can be recognized. In this paper, we analyze how this limitation affects performance by comparing the proposed method with existing CL approaches.

To this end, we adapt four representative CL methods to be compatible with the OVS model. Following prior works~\citep{surveycontinual, chen2022lifelong, parisi2019continual, mundt2023wholistic}, we categorize existing CL methods into three groups: replay-based, regularization-based (parameter, function), and parameter-isolation-based. We select representative methods from each category and apply it to the OVS model. Section~\ref{sec:replayCL} to \ref{sec:parameterCL} describes the overview of each method and the modifications made to align them with the OVS setting.

\subsection{Replay-based Method: ER}
\label{sec:replayCL}
Experience Replay (ER) stores a fixed number of samples per class from previous training datasets and uses them when training on new datasets. This technique serves as the conceptual foundation for various memory-based continual learning methods~\citep{GEM, iscen2020memory}.

To apply ER to the OVS model, we select ten samples per class from the previous training dataset. Since OVS uses a multi-label structure where each image can contain multiple classes, we prevent duplicate selection by sequentially sampling ten images per class without redundancy.

ER assumes access to the pre-training dataset during the training of new datasets. Therefore, there is a limitation in conducting a fair comparison with the proposed method or other CL methods that do not require access to pre-training data. Nevertheless, we include ER in our comparison to provide a broad performance analysis across different CL strategies.

\subsection{Regularization-based Methods: LwF \& EWC}
\label{sec:regularizationCL}
Regularization-based methods include two subtypes: (1) function regularization and (2) parameter regularization. For function regularization, we apply Learning without Forgetting (LwF)~\citep{lwf} to the OVS model. LwF uses a knowledge distillation loss based on the prediction distance between the previously trained model and the newly trained model. Since LwF constructs its loss using the probability scores generated by the previously trained model, it requires the preservation of a classification head. However, OVS models do not include classification heads. To address this, we instead compute the similarity between the text embedding of the previously trained model and the class embeddings to obtain probability scores used for distillation.

Second, we apply Elastic Weight Consolidation (EWC)~\citep{ewc}, a parameter regularization method, to the OVS model. EWC estimates the Fisher Information Matrix using the new training dataset, which measures the importance of each parameter. It then penalizes changes to important parameters, thereby preserving previously learned knowledge. Since this method does not alter the architecture of the OVS model, it can be applied without structural modifications. However, the effectiveness of this approach depends on clear separation between datasets. If there are overlapping or similar classes or domains between the datasets, integrating knowledge from both can improve performance. EWC, however, restricts updates to parameters deemed important for the first dataset, which hinders the model’s ability to incorporate new knowledge from the second dataset. As a result, even when the datasets share useful information, the model cannot integrate it effectively, making EWC less suitable for scenarios involving sequential dataset training.

\subsection{Parameter-isolation-based Method: ECLIPSE}
\label{sec:parameterCL}
We apply ECLIPSE~\citep{eclipse}, a parameter-isolation-based method, to the OVS model. ECLIPSE is designed for class-incremental learning in closed-set segmentation settings. Specifically, it adds learnable prompts to the object queries and positional embeddings when learning new classes, and updates the classification head accordingly. Since OVS models do not use a classification head, we exclude the classification head component and apply only the prompt tuning elements.

ECLIPSE also utilizes logit manipulation based on class-wise probability scores obtained from the classification head. This component helps determine whether the predicted mask from an object query corresponds to a valid class or to a “no object” case, preventing semantic drift for unseen classes in closed-set settings. However, the OVS model must recognize unseen classes and does not include a classification head, which makes direct application of the logit manipulation method infeasible. Therefore, we also exclude the logit manipulation component of ECLIPSE in our implementation.

Lastly, because each incremental dataset contains a large number of classes, we assume that a sufficient number of learnable parameters is necessary. Accordingly, we configure the model to learn 250 additional prompts per dataset.

\subsection{Alternative Approaches to MVN Distribution}
\label{app:comparison_description_mvn}

\myparagraph{K-Means Clustering} is an unsupervised learning algorithm that partitions a given dataset into K clusters. Each cluster is associated with a centroid, and the algorithm iteratively assigns each data point to the cluster whose centroid is closest, then updates the centroid as the mean of all data points assigned to the cluster. Clustering is based on the Euclidean distance, and the algorithm aims to minimize the sum of squared distances between data points and their corresponding centroids. K-Means is computationally efficient and easy to implement. However, it struggles with robustness under noisy conditions and cannot effectively model non-spherical or overlapping distributions.

\myparagraph{Kernel Density Estimation (KDE)} is a non-parametric method for estimating the probability distribution of a given dataset. KDE constructs the probability density function by placing a kernel function on each data point and summing their contributions. In addition, a bandwidth parameter in KDE controls the smoothness of the resulting density: small bandwidths lead to overfitting, while large bandwidths oversmooth the distribution. For that reason, we set the bandwidth to 0.5. Due to its flexibility, KDE can approximate arbitrary distributions without requiring any parametric assumptions. However, unlike the MVN distribution, KDE does not offer a compact parametric form. As a result, it suffers from unstable likelihood estimation, making it less suitable for tasks that rely on density-based inference, such as interpolation factor estimation.

\section{Details of Task Vectors}

\begin{figure}[ht]
    \centering
    \includegraphics[width=0.5\textwidth]{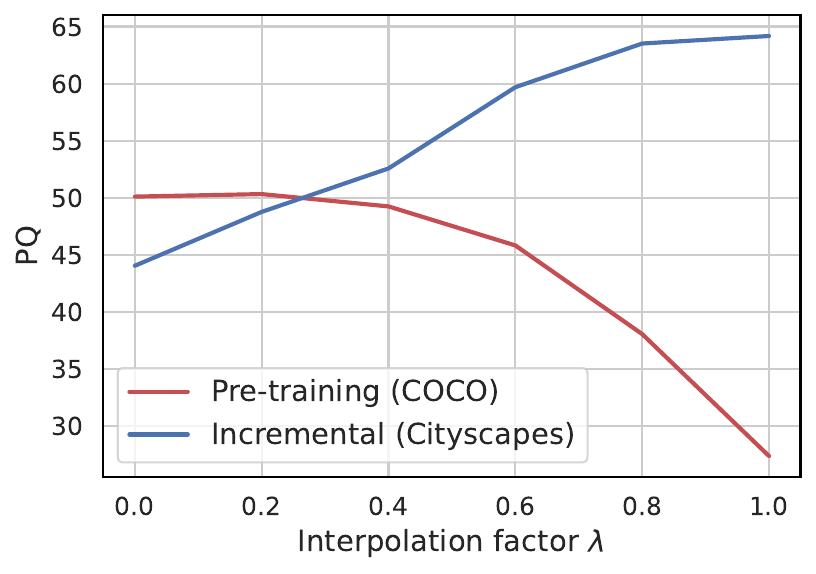}
    \caption{\small
    Performance on the evaluation set of Cityscapes and COCO depending on the interpolation factor $\lambda$, using fc-clip.
    }
    \label{fig:interpolation_factor}
\end{figure}

The task vector~\citep{ilharco2022editing} is constructed by subtracting the weights of a pre-trained model from those of a model fine-tuned on a specific task. These task vectors can be modified or combined through arithmetic operations such as addition, and the behavior of the resulting model is adjusted accordingly. For example, by adding two task vectors obtained from fine-tuning on different tasks and then adding the result to the pre-trained model's weights, one can generate model weights that can be utilized on both tasks. In addition, an interpolation factor $\lambda$ can be multiplied by the task vector. This $\lambda$ determines whether the model uses the weights trained on the pre-training dataset or those trained on the fine-tuning dataset. 

To verify whether the combination of interpolation factors and task vectors is also effective in OVS models, we construct a task vector by subtracting the weights of a pre-trained OVS model from those of a fine-tuned OVS model and evaluate the performance changes when various $\lambda$ values are multiplied to this vector. As shown in Figure~\ref{fig:interpolation_factor}, when $\lambda = 0$, the decoder uses the pre-trained weights $\theta_{\text{dec,pr}}$, which results in strong performance on the pre-training dataset. In contrast, when $\lambda$ is close to 1, the decoder uses weights close to the fine-tuned weights $\theta_{\text{dec,ft}}$, leading to high performance on the fine-tuning dataset. When $\lambda$ takes a value between 0 and 1, the decoder interpolates between the two weights and achieves balanced performance across both datasets.

\section{Computational Resources}

\subsection{Training Resources}
\label{app:training_resources}

\begin{table}[ht]
\centering
\scriptsize
\setlength{\tabcolsep}{1.2em}
\renewcommand{\arraystretch}{1.1}
\caption{Comparison of training time and GPU memory usage across different methods. The values are reported relative to standard fine-tuning. For reference, fine-tuning required 5110 seconds and 22.7 GB in Scenario 1, and 5701 seconds and 20.8 GB in Scenario 2.}
\begin{tabular}{l cc cc}
\toprule
\multirow{2}{*}{\textbf{Method}} & \multicolumn{2}{c}{\textbf{Scenario 1 (Cityscapes)}} & \multicolumn{2}{c}{\textbf{Scenario 2 (ADE20K)}} \\
\cmidrule(lr){2-3} \cmidrule(lr){4-5}
 & \textbf{Training Time} & \textbf{GPU Memory} & \textbf{Training Time} & \textbf{GPU Memory} \\
\midrule
Fine-tuning    & 1.00 & 1.00 & 1.00 & 1.00 \\
Retraining     & 1.25 & 0.86 & 1.40 & 0.95 \\
ER             & 1.25 & 0.86 & 1.40 & 0.95 \\
LwF            & 1.28 & 1.07 & 1.21 & 1.13 \\
EWC            & 1.07 & 1.00 & 1.01 & 1.03 \\
ECLIPSE        & 0.75 & 0.57 & 0.74 & 0.52 \\
ConOVS (ours)  & 1.00 & 1.00 & 1.00 & 1.00 \\
\bottomrule
\end{tabular}
\label{tab:training_cost}
\end{table}

To provide a fair comparison, we measured the training time and GPU memory usage of each method and normalized all results to the computational cost of fine-tuning. As shown in Table~\ref{tab:training_cost}, training time does not vary substantially across methods, which is expected since the number of training iterations was fixed for all experiments. An exception is ECLIPSE, which demonstrates lower training cost due to its use of Visual Prompt Tuning, updating only a small subset of model parameters. While this design improves training efficiency, our results show that ConOVS achieves superior segmentation performance compared to ECLIPSE. We attribute this performance gap to the inherent limitations of methods that update only a fraction of the model parameters.

\subsection{Inference Resources}

\begin{figure}[ht]
	\centering
	\scriptsize
	\setlength{\columnsep}{10pt}
	\begin{subfigure}[b]{0.49\linewidth}
		\centering
		\includegraphics[width=0.99\linewidth]{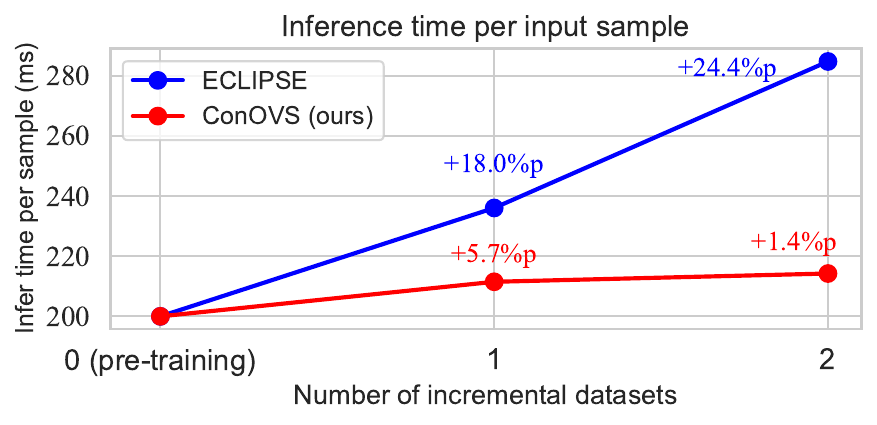}
		\caption{\small Inference time}
		\label{fig:inf_time_comparison}
	\end{subfigure}
	\hspace{0.5em}
	\begin{subfigure}[b]{0.49\linewidth}
		\centering
		\includegraphics[width=0.99\linewidth]{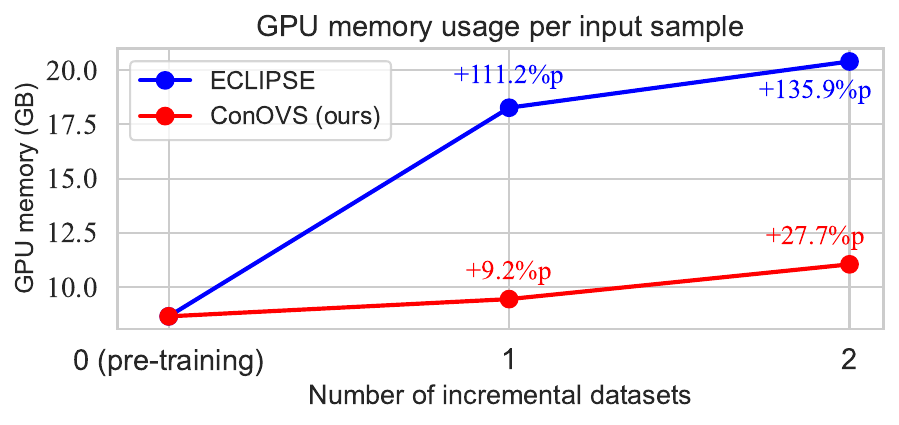}
		\caption{\small GPU memory}
		\label{fig:memory_comparison}
	\end{subfigure}
	\caption{\small Comparison of per-sample inference resources between ConOVS (ours) and ECLIPSE.}
	\label{fig:resource_comparison}
\end{figure}

Our method is based on a Mixture-of-Experts (MoE) framework that dynamically combines multiple expert models according to the input sample. In MoE-based continual learning~\citep{le2024mixture}, a new expert is added each time an incremental dataset is introduced. This raises concerns that the number of parameters involved in inference may increase with the number of incremental datasets, potentially leading to higher inference time and GPU memory usage~\citep{huang2024toward}. To examine this issue, we measure how resource usage scales with the number of incremental datasets and compare our method against ECLIPSE~\citep{eclipse}, a parameter-isolation-based method that adds parameters during incremental learning.

As shown in Figure~\ref{fig:resource_comparison}, our method consistently requires less inference time and GPU memory than ECLIPSE. In addition, as the number of incremental datasets increases, the growth in inference time is significantly smaller for our method. This difference arises from how the additional parameters are utilized during inference. In ECLIPSE, all parameters added during training are used at inference, so the number of parameters used increases in proportion to the number of incremental datasets. In contrast, our method interpolates expert weights based on the input sample to generate a single decoder weight, so the number of parameters used during inference remains constant regardless of the number of trained datasets.

\fixall{
\begin{table*}[h]
\centering
\vspace{-.5em}
\scriptsize
\setlength{\tabcolsep}{1.2em}
\renewcommand{\arraystretch}{1.1}
\caption{\small Inference time per sample with varying numbers of incremental datasets. The unit for all numbers in the table is milliseconds (ms).}

\begin{tabular}{@{}ccccccc@{}}
\toprule
\makecell{\textbf{Number of} \\ \textbf{Incremental Datasets}} & \textbf{Encoder} & \makecell{\textbf{Interpolation Factor} \\ \textbf{Estimator}} & \makecell{\textbf{Expert} \\ \textbf{Interpolation}} & \textbf{Decoder} & \makecell{\textbf{Total Inference} \\ \textbf{Time Per Sample}} & \textbf{Change (\%)} \\ \midrule
0                                & 97.69                & -                                           & -                                         & 102.30              & 199.99                                   & +0.00\%             \\
1                                & 97.69                & 0.81                                        & 10.69                                     & 102.30              & 211.48                                   & +5.75\%             \\
2                                & 97.69                & 1.01                                        & 13.23                                     & 102.30              & 214.23                                   & +7.12\%             \\ \bottomrule
\end{tabular}

\label{tab:inference_time}
\end{table*}
}

In practice, our method only incurs additional resource usage when estimating the interpolation factor or interpolating the expert weights; otherwise, the resource usage of the model itself remains unchanged. As shown in Table~\ref{tab:inference_time}, the inference time of the encoder and decoder parts does not increase even as the number of incremental datasets grows.

Beyond inference time, our method also achieves high efficiency in terms of storage. Unlike ensemble-based approaches~\citep{modelsoup, khirodkar2022sequential}, which require storing the full model weights to combine multiple models, our method stores only the parameters of the decoder. As a result, it is sufficient to store only $6.11\%$ of the total model size, which corresponds to approximately 80MB per dataset. This efficiency ensures high scalability even in scenarios where the number of models to be combined gradually increases.

\section{More Incremental Datasets}
\label{app:more_incremental}

\begin{table*}[ht]
\centering
\scriptsize
\setlength{\tabcolsep}{0.4em}
\renewcommand{\arraystretch}{1.1}
\caption{\small Performance comparison across pre-training, incremental, and zero-shot datasets when sequentially training on five datasets. The best performance for each dataset is highlighted in bold.}
\begin{tabular}{@{}l ccccc cccccc c@{}}
\toprule
\multirow{2}{*}{\textbf{Method}} & 
\multicolumn{1}{c}{\textbf{Pre-training}} &
\multicolumn{4}{c}{\textbf{Incremental}} & 
\multicolumn{6}{c}{\textbf{Zero-shot}} & 
\multirow{2}{*}{\textbf{Average}} \\
\cmidrule(lr){2-2} \cmidrule(lr){3-6} \cmidrule(lr){7-12}
& COCO & Cityscapes & A-150 & BDD100K & Mapillary & LVIS & PC-59 & PC-459 & PAS-20 & PAS-21 & A-847 & \\
\midrule \midrule
fc-clip       & 50.1 & 44.0 & 23.5 & 19.0 & 26.0 & 20.5 & 53.0 & 16.9 & 93.1 & 80.2 & 13.8 & 40.0 \\
Fine-tuning   & 31.7 & 55.5 & 28.7 & 25.8 & 36.3 & 17.7 & 49.8 & 16.3 & 90.0 & 73.9 & 14.0 & 40.0 \\
Retraining    & 49.0 & 62.3 & 36.3 & 28.2 & 34.5 & 21.7 & \textbf{53.6} & \textbf{17.4} & \textbf{94.0} & \textbf{80.7} & 15.6 & 44.8 \\
\rowcolor[HTML]{EFEFEF}
\textbf{ConOVS (ours)} & \textbf{50.1} & \textbf{62.9} & \textbf{38.5} & \textbf{29.1} & \textbf{35.0} & \textbf{21.8} & \textbf{53.6} & 17.3 & 93.2 & 80.4 & \textbf{16.0} & \textbf{45.3} \\
\bottomrule
\end{tabular}
\label{tab:more_incremental}
\end{table*}

To validate the effectiveness of our method on a larger number of incremental datasets, we designed Scenario 4 (\textbf{S4}) in which the model is sequentially trained on five datasets: COCO, Cityscapes, ADE20K, BDD100K, and Mapillary Vistas. Although we aimed to include additional datasets, we were constrained to those that provide panoptic segmentation annotations.

The results of this extended experiment are presented in Table~\ref{tab:more_incremental}. As shown, our method outperforms the baseline, fine-tuning, and retraining approaches. These findings demonstrate that our method remains effective as the number of incremental datasets increases, highlighting its scalability under complex domain shifts.

\section{Additional Ablation Studies of ConOVS}
\label{app:additional_ablation}

\subsection{Ablation Study on Fine-tuned Components}

\begin{table}[ht]
\centering
\scriptsize
\caption{Ablation study on different fine-tuned components. The best performance for each dataset is highlighted in bold.}
\begin{tabular}{lccc}
\toprule
\textbf{Fine-tuned Component} & \textbf{COCO (pre-training)} & \textbf{Cityscapes (incremental)} & \textbf{ADE20K (zero-shot)} \\
\midrule
fc-clip       & 50.1 & 44.0 & 23.5 \\
Last Block    & 50.0 & 50.4 & 25.8 \\
LayerNorm     & 49.9 & 44.0 & 25.8 \\
LoRA          & \textbf{50.7} & 47.9 & 25.9 \\
Only decoder  & 50.4 & \textbf{64.4} & \textbf{26.0} \\
\bottomrule
\end{tabular}
\label{tab:ablate_finetuning}
\end{table}

Given recent works that fine-tune the CLIP encoder~\cite{catseg, lee2025effective}, a more detailed ablation study on our design choice to freeze the encoder and fine-tune only the decoder is necessary. To this end, we experimented with three encoder fine-tuning strategies: (1) fine-tuning only the last block of the encoder, (2) fine-tuning only the LayerNorm modules, and (3) replacing all MLP layers with LoRA modules and fine-tuning them. Note that, unlike prior attention-based methods designed for ViT-style CLIP encoders~\cite{catseg, lee2025effective}, our strategies are tailored to the ConvNeXt-based encoders used in fc-clip.

As shown in Table~\ref{tab:ablate_finetuning}, all three encoder fine-tuning strategies yield lower performance on the incremental dataset compared to our design choice of fine-tuning only the decoder. We attribute this to an architectural difference: whereas previous methods~\cite{catseg, lee2025effective} generate segmentation masks directly from the encoder and thus benefit from encoder fine-tuning, fc-clip generates masks in the decoder's mask head. This suggests that decoder fine-tuning is more effective for improving segmentation performance in our setup.

\subsection{Ablation Study of Softmax Operation}

\begin{table}[ht]
\centering
\scriptsize
\caption{Performance comparison with different normalization operations. The best results for each column are highlighted in bold.}
\begin{tabular}{lcccc}
\toprule
\textbf{Operation} & \textbf{COCO (pretraining)} & \textbf{Cityscapes (incremental)} & \textbf{ADE20K (zero-shot)} & \textbf{Average} \\
\midrule
minmax   & 50.3 & 50.5 & \textbf{26.1} & 42.3 \\
sigmoid  & 50.3 & 50.3 & 26.0 & 42.2 \\
softmax  & \textbf{50.4} & \textbf{64.4} & 26.0 & \textbf{46.9} \\
\bottomrule
\end{tabular}
\label{tab:ablate_softmax}
\end{table}

We apply the softmax operation to the log-likelihood vector to normalize the proximity scores of each domain to the $[0, 1]$ range. This decision is based on a prior study~\cite{ilharco2022editing}, which found that when the interpolation factor exceeds 1, merging performance degrades. To normalize the log-likelihood values obtained from the MVN distribution, we considered three strategies, including min-max normalization, sigmoid, and softmax. As shown in Table~\ref{tab:ablate_softmax}, softmax achieved the best performance, which led us to adopt it.

\subsection{Ablation Study of Element-wise Maximum Operation}

\begin{table}[ht]
\centering
\scriptsize
\caption{Comparison of operations for calculating the interpolation factor $\lambda$, with the best result in each column highlighted in bold.}
\begin{tabular}{lccccc}
\toprule
\textbf{Input} & \textbf{Operation} & \textbf{COCO (pre-training)} & \textbf{Cityscapes (incremental)} & \textbf{ADE20K (zero-shot)} & \textbf{Average} \\
\midrule
image only & -            & 51.5 & 43.4 & 25.8 & 40.2 \\
text only  & -            & \textbf{51.9} & 60.7 & 25.9 & 46.2 \\
image+text & Average      & 50.3 & 64.2 & 25.9 & 46.8 \\
image+text & Multiplication & 50.1 & 63.8 & 25.8 & 46.6 \\
image+text & Maximum      & 50.4 & \textbf{64.4} & \textbf{26.0} & \textbf{46.9} \\
\bottomrule
\end{tabular}
\label{tab:ablate_maximum}
\end{table}

We use the element-wise maximum operation to combine information from both the image and text modalities. Initially, we considered whether to rely on only the image domain, only the text domain, or both. For combining both modalities, we evaluated three options: average, multiplication, and element-wise maximum. 

As shown in Table~\ref{tab:ablate_maximum}, the results show that combining both modalities performs better than using a single modality, and among the combination methods, element-wise maximum achieved the best performance. Based on these findings, we selected element-wise maximum as our fusion strategy.

\section{Additional Analysis}

\subsection{Evaluation on Diverse and Challenging Domains}

\begin{table}[ht]
\centering
\scriptsize
\setlength{\tabcolsep}{.7em}
\renewcommand{\arraystretch}{1.3}
\caption{Performance comparison (\miou) on datasets with significant domain shifts.}
\scriptsize
\begin{tabular}{lccc}
\hline
\textbf{Method}      & \textbf{GTA5} & \textbf{DarkZurich} & \textbf{FoggyZurich} \\ \hline \hline
fc-clip     & 65.6          & 40.2                & 54.4                \\
+ Fine-tuning & 58.4        & 39.8                & 52.1                \\
\rowcolor[HTML]{EFEFEF} \textbf{+ ConOVS (ours)}      & \textbf{66.6} & \textbf{43.1}       & \textbf{55.9}       \\
\end{tabular}
\label{tab:adverse_synthetic}
\end{table}

To demonstrate the robust generalization capability of the proposed method, we evaluate the model on three zero-shot test datasets that differ significantly from the domains of the training datasets. Specifically, we train the model using COCO as the pre-training dataset and ADE20K as the incremental dataset. For evaluation, we use GTA5~\citep{gtav} (a synthetic driving simulation), DarkZurich~\citep{darkzurich} (nighttime driving scenes), and FoggyZurich~\citep{foggyzurich} (driving scenes with fog), which represent domains that are substantially different from the training data.

As shown in Table~\ref{tab:adverse_synthetic}, simply fine-tuning fc-clip on the incremental dataset leads to performance degradation across all three zero-shot test datasets. In contrast, the proposed method improves performance on all of them, demonstrating its effectiveness even under adverse conditions and in synthetic environments with large domain shifts.

\subsection{Different Pre-training Dataset}

\begin{table*}[ht]
    \centering
    \caption{\small \fix{Performance comparison among fc-clip, fine-tuning, and ConOVS (ours), with ADE20K as the pre-training dataset. The incremental dataset is either (a) COCO or (b) Cityscapes. \pq is used.}}
    \fix{
    \begin{minipage}{0.47\textwidth}
        \centering
        \scriptsize
\setlength{\tabcolsep}{0.5em}
\renewcommand{\arraystretch}{1.0}
\begin{tabular}{@{}lc|c|c@{}}
\textbf{Method} & \textbf{\begin{tabular}[c]{@{}c@{}}ADE20K \\ (\pretraining)\end{tabular}} & \textbf{\begin{tabular}[c]{@{}c@{}}COCO \\ (incremental)\end{tabular}} & \textbf{\begin{tabular}[c]{@{}c@{}}Cityscapes \\ (\unseen)\end{tabular}} \\ \midrule \midrule
fc-clip & 48.1 & 42.3 & 40.9 \\ \midrule
Fine-tuning & \tableminus{-18.5} & \textbf{\tableplus{+10.4}} & \tableplus{+3.3} \\
\rowcolor[HTML]{EFEFEF}
\textbf{ConOVS (ours)} & \textbf{\tableminus{-1.3}} & \tableplus{+9.3} & \textbf{\tableplus{+5.2}}
\end{tabular}
        \subcaption{\small COCO\label{tab:ade_coco}}
    \end{minipage}
    \hfill
    \begin{minipage}{0.47\textwidth}
        \centering
        \scriptsize
\setlength{\tabcolsep}{.5em}
\renewcommand{\arraystretch}{1.0}
\begin{tabular}{@{}lc|c|c@{}}
\textbf{Method} & \textbf{\begin{tabular}[c]{@{}c@{}}ADE20K \\ (\pretraining)\end{tabular}} & \textbf{\begin{tabular}[c]{@{}c@{}}Cityscapes \\ (incremental)\end{tabular}} & \textbf{\begin{tabular}[c]{@{}c@{}}COCO \\ (\unseen)\end{tabular}} \\ \midrule \midrule
fc-clip & 48.1 & 40.9 & 42.3 \\ \midrule
Fine-tuning & \tableminus{-18.5} & \textbf{\tableplus{+21.4}} & \tableminus{-11.5} \\
\rowcolor[HTML]{EFEFEF}
\textbf{ConOVS (ours)} & \textbf{+0.0} & \tableplus{+19.5} & \textbf{+0.0}
\end{tabular}
        \subcaption{\small Cityscapes\label{tab:ade_cityscapes}}
    \end{minipage}
    }
\label{tab:single_ade}
\end{table*}

To verify the effectiveness of the proposed method on OVS models pre-trained on datasets other than COCO, we apply our method to an OVS model pre-trained on ADE20K and compare the performance. As shown in Table~\ref{tab:single_ade}, even when the pre-training dataset changes, the proposed method significantly improves performance on the incremental dataset compared to the baseline fc-clip (e.g., +19.5 on Cityscapes). Moreover, compared to fine-tuning, the proposed method maintains the performance on the pre-training dataset (e.g., on ADE20K, Fine-tuning: -18.5, ConOVS: +0.0) while achieving strong performance on the incremental dataset. These results suggest that the proposed method can effectively expand the recognition capability of OVS models by learning newly collected datasets, regardless of the type of pre-training dataset.

\subsection{Performance of Retraining Across Training Iterations}
\label{app:iter_retraining}

\begin{table*}[ht]
\centering
\scriptsize
\caption{Performance comparison across training iterations of the retraining. \pq is used.}
\begin{tabular}{l c c c c}
\toprule
\textbf{Method} & \textbf{Training Iterations} & \textbf{COCO (pre-training)} & \textbf{Cityscapes (incremental)} & \textbf{ADE20K (zero-shot)} \\
\midrule
Retraining & 10k  & 50.7 & 61.9 & 25.2 \\
Retraining & 20k  & 50.7 & 62.5 & 25.4 \\
Retraining & 40k  & \textbf{51.0} & 63.3 & 25.3 \\
Retraining & 100k & \textbf{51.0} & \textbf{64.4} & 25.5 \\
\rowcolor[HTML]{EFEFEF}
ConOVS (ours) & 10k  & 50.4 & 64.2 & \textbf{26.0} \\
\bottomrule
\end{tabular}
\label{tab:iter_retraining}
\end{table*}

As shown in Section~\ref{sec:experiments}, the retraining method demonstrates relatively lower performance. This is because all methods were conducted under the same computational budget. Since retraining requires learning from a substantially larger amount of data compared to our method, it needs a longer training schedule to reach convergence.

To analyze this more precisely, we conducted additional experiments by training the retraining method for longer durations. As presented in Table~\ref{tab:iter_retraining}, when trained with a sufficiently long schedule (100k iterations), the retraining method achieves better performance than our approach on the pretraining and incremental datasets. However, despite consuming significantly more computational resources, retraining does not provide a substantial improvement over ConOVS. One possible explanation for this limited performance is Task Interference~\cite{gradientsurgery, gradnorm}, which may occur when training on multiple domains simultaneously. In contrast, ConOVS avoids this issue by independently training domain-specific experts and dynamically combining them at inference time. Consequently, our method achieves performance comparable to retraining while requiring significantly less computational cost.

\subsection{Effectiveness of ConOVS with Multi-Domain Incremental Datasets}

\begin{table}[ht]
\centering
\scriptsize
\setlength{\tabcolsep}{0.5em}
\renewcommand{\arraystretch}{1.0}
\caption{Comparison between the baseline and ConOVS when the incremental dataset consists of multiple domains.}
\begin{tabular}{l c c c c}
\toprule
\textbf{Method} & \textbf{\begin{tabular}[c]{@{}c@{}}COCO\\(pre-training)\end{tabular}} & \textbf{\begin{tabular}[c]{@{}c@{}}Cityscapes\\(incremental)\end{tabular}} & \textbf{\begin{tabular}[c]{@{}c@{}}ADE20K\\(incremental)\end{tabular}} & \textbf{\begin{tabular}[c]{@{}c@{}}Average on \\4 zero-shot datasets\end{tabular}} \\
\midrule
fc-clip        & 50.1 & 44.0 & 23.5 & 60.8 \\
\rowcolor[HTML]{EFEFEF}
ConOVS (ours)  & \textbf{50.1} & \textbf{60.4} & \textbf{31.5} & \textbf{61.3} \\
\bottomrule
\end{tabular}
\label{tab:multiple_domain_incremental}
\end{table}

We conducted an additional experiment to assess whether our method remains effective when a single incremental dataset contains samples from multiple domains. For this purpose, we combined Cityscapes and ADE20K into a single incremental dataset. Note that the performance on the zero-shot setting was measured by averaging results across four datasets: PC-59, PC-459, PAS-20, and PAS-21.

As shown in Table~\ref{tab:multiple_domain_incremental}, our method maintains the performance of the baseline fc-clip on the pre-training dataset, while significantly improving results on both the incremental and zero-shot datasets. These findings demonstrate that ConOVS remains robust and effective even when the incremental dataset spans diverse domains.

\subsection{Effectiveness of ConOVS in Low-Resource Incremental Settings}

\begin{table}[ht]
\centering
\scriptsize
\caption{Performance comparison between fc-clip and ConOVS with a small incremental dataset (5\% of Cityscapes). ConOVS maintains pre-training performance while improving incremental and zero-shot results, showing robustness in low-resource settings.}
\begin{tabular}{l c c c}
\toprule
\textbf{Method} & \textbf{COCO (pre-training)} & \textbf{Cityscapes (incremental)} & \textbf{ADE20K (zero-shot)} \\
\midrule
fc-clip        & 50.1 & 44.0 & 23.5 \\
\rowcolor[HTML]{EFEFEF}
ConOVS (ours)  & 50.0 & \textbf{48.8} & \textbf{25.9} \\
\bottomrule
\end{tabular}
\label{tab:fewshot}
\end{table}

To investigate whether the proposed method remains effective under low-resource conditions, we conducted an additional experiment using a small incremental dataset. Specifically, we randomly sampled 5\% of the Cityscapes training set and used it as the incremental dataset.

As shown in Table~\ref{tab:fewshot}, our method maintains performance on the pre-training dataset while improving results on both the incremental and zero-shot datasets, even with the reduced incremental data. These findings demonstrate that ConOVS remains effective in scenarios where the number of incremental samples is limited.

\subsection{Comparison with SemLA}

Compared to SemLA~\cite{semla}, our approach shares a conceptual similarity in that both methods compute the proximity of an input image to each training dataset and use it to determine dynamic merging weights during inference. However, our method differs in three key design choices: (1) it uses both image and text embeddings to assess domain relevance, whereas SemLA relies solely on image embeddings; (2) it estimates domain proximity using multivariate normal (MVN) distributions, while SemLA computes L2 distances to dataset-specific centroids; and (3) it fine-tunes the decoder for each dataset, in contrast to SemLA, which applies LoRA modules to the CLIP image encoder. 

To provide a direct comparison, we re-implemented SemLA within our experimental framework. In this variant, domain proximity was computed only from image embeddings, estimated by calculating the L2 distance between the input embedding and dataset-specific centroids, and model adaptation was carried out through LoRA fine-tuning on the CLIP image encoder instead of decoder fine-tuning. 

\begin{table}[ht]
\centering
\scriptsize
\caption{Performance comparison between fc-clip, SemLA, and ConOVS. The best performance for each dataset is highlighted in bold.}
\begin{tabular}{lccc}
\toprule
\textbf{Method} & \textbf{COCO (pre-training)} & \textbf{Cityscapes (incremental)} & \textbf{ADE20K (zero-shot)} \\
\midrule
fc-clip        & 50.1 & 44.0 & 23.5 \\
SemLA          & \textbf{50.9} & 61.1 & \textbf{26.0} \\
ConOVS (ours)  & 50.4 & \textbf{64.4} & \textbf{26.0} \\
\bottomrule
\end{tabular}
\label{tab:comparison_semla}
\end{table}

As shown in Table~\ref{tab:comparison_semla}, our method (ConOVS) outperforms this SemLA variant on the incremental dataset. This improvement suggests that our design choices are better suited for continual open-vocabulary segmentation.

\section{Discussion}

\myparagraph{Applicability to Open-Vocabulary Object Detection (OVD).} Our proposed method is not limited to Open-Vocabulary Segmentation (OVS) but can also be extended to OVD frameworks. Specifically, our approach is applicable to models with an encoder-decoder architecture that perform classification based on the similarity between image and text embeddings. This structural characteristic is shared by most OVD methods. For instance, YOLO-World~\cite{yoloworld} adopts a modular design consisting of an encoder and a prediction head, and omits the use of a conventional fc-based classifier. Given this structure, all components of our method can be directly incorporated without modification. 

To be more specific, for each newly introduced training dataset, an MVN distribution can be formed using the encoder's embeddings. During inference, the interpolation weights can be dynamically adjusted based on the proximity between the input sample and each domain, enabling the model to incrementally extend its recognition capability.

\myparagraph{Effectiveness in Weakly Supervised Settings.} We believe that ConOVS can function effectively regardless of the issue of low-quality segmentation annotations in weakly labeled datasets~\cite{hwang2023entropy, hwang2025curriculum, hwang2025small}. This is because our technique constructs the MVN distribution using only image and text embeddings, without relying on segmentation annotations. Therefore, we see no constraints in forming the MVN distribution and expect to accurately estimate the interpolation coefficients. While label noise may affect the overall performance, we contend that our method can still contribute to effectively enhancing the model’s recognition ability even in weakly labeled environments.

\myparagraph{Applicability to Cost-based OVS Methods.} Our method is also applicable to cost-based OVS approaches~\cite{catseg}. Since it is designed to merge independently trained models, it remains compatible with techniques that fine-tune the CLIP encoder. While cost-based methods typically generate segmentation maps by post-processing encoder features, our method does not rely on such steps, making it directly applicable without modification.

However, when applied to cost-based OVS models, the encoding process must be executed twice. This is because our method computes the interpolation factor based on the proximity of the input sample, which requires an initial forward pass through the encoder. After the interpolation factor is determined, a second forward pass is performed using the merged encoder to generate the final feature representation. Consequently, an increase in inference time is expected.

\section{Qualitative Results}
\label{app:qualitative}

\begin{figure}[ht]
    \centering
    \includegraphics[width=0.93\linewidth]{figure/qualitative.pdf}
    \caption{\small
    We provide a qualitative analysis on COCO, Cityscapes, and ADE20K. The comparison includes three methods: fc-clip, fine-tuning, and ConOVS (ours). Cityscapes is used as the incremental dataset.
    }
    \label{fig:qualitative}
\end{figure}

This section presents a qualitative analysis of the original fc-clip, the standard fine-tuning technique, and ConOVS (ours). Figure~\ref{fig:qualitative} illustrates the qualitative outputs of each method. On the pre-training dataset, the fine-tuning technique fails to recognize the \textit{backpack}, indicating a loss of information from the pre-training stage. On the incremental dataset, fc-clip fails to identify key elements such as \textit{road} and \textit{person}, highlighting that OVS models perform well only within the distribution of the pre-training dataset. On the zero-shot dataset, the fine-tuning technique fails to recognize \textit{ceiling}, a class absent from both the pre-training and incremental dataset. In contrast, the proposed method successfully identifies both previously learned and newly introduced classes, as well as classes not included in either training dataset.

\section{Results of All Evaluation Metrics}
\label{app:all_quantitative}

In this study, we consider fc-clip and X-Decoder as baseline models, both of which are designed to handle panoptic, instance, and semantic segmentation tasks. Based on this capability, we assess their performance using the corresponding metrics: \pq, \map, and \miou. As shown in Tables~\ref{tab:cityscapes_all} to~\ref{tab:multiple_all}, the proposed method consistently outperforms all other approaches across the pre-training, incremental, and zero-shot test datasets, regardless of the evaluation metric. This confirms, as noted in Section~\ref{sec:eval_metrics}, that the three metrics follow similar performance trends. Moreover, the results verify that the proposed method performs robustly across all segmentation tasks.

\clearpage

\begin{table}[h]
\scriptsize
\setlength{\tabcolsep}{0.7em}
\renewcommand{\arraystretch}{1.2}
\caption{\small Performance comparison among the baseline, fine-tuning, retraining, existing continual learning methods, and ConOVS (ours). The incremental dataset is Cityscapes. We use \pq, \map, and \miou as evaluation metrics.}
\begin{tabular}{@{}lccccc|cccc|cccc@{}}
 &  & \multicolumn{4}{c|}{\textbf{COCO (pre-training)}} & \multicolumn{4}{c|}{\textbf{Cityscapes (incremental)}} & \multicolumn{4}{c}{\textbf{ADE20K (zero-shot)}} \\
\multirow{-2}{*}{\textbf{Method}} & \multirow{-2}{*}{\textbf{\begin{tabular}[c]{@{}c@{}}Incremental \\ Dataset\end{tabular}}} & \textbf{\pq} & \textbf{\map} & \textbf{\miou} & \cellcolor[HTML]{EFEFEF}\textbf{\texttt{Avg}} & \textbf{\pq} & \textbf{\map} & \textbf{\miou} & \cellcolor[HTML]{EFEFEF}\textbf{\texttt{Avg}} & \textbf{\pq} & \textbf{\map} & \textbf{\miou} & \cellcolor[HTML]{EFEFEF}\textbf{\texttt{Avg}} \\ \Xhline{3\arrayrulewidth}
\addlinespace[3pt]
fc-clip & - & 50.1 & 41.1 & 52.0 & \cellcolor[HTML]{EFEFEF}47.7 & 44.0 & 26.8 & 56.2 & \cellcolor[HTML]{EFEFEF}42.4 & 23.5 & 17.1 & 30.4 & \cellcolor[HTML]{EFEFEF}23.7 \\ \midrule
Fine-tuning &  & \tableminus{\tableminus{-22.7}} & \tableminus{\tableminus{-16.2}} & \tableminus{\tableminus{-11.8}} & \cellcolor[HTML]{EFEFEF}\tableminus{\tableminus{-16.9}} & \tableplus{+20.1} & \tableplus{+13.9} & \tableplus{+21.2} & \cellcolor[HTML]{EFEFEF}\tableplus{+18.4} & \tableminus{-10.3} & \tableminus{-6.3} & \tableminus{-3.9} & \cellcolor[HTML]{EFEFEF}\tableminus{-6.8} \\
Retraining &  & \tableplus{+0.6} & \tableplus{+0.9} & \tableplus{+0.3} & \cellcolor[HTML]{EFEFEF}\tableplus{+0.6} & \tableplus{+17.9} & \tableplus{+10.4} & \tableplus{+18.7} & \cellcolor[HTML]{EFEFEF}\tableplus{+15.7} & \tableplus{+1.7} & \tableminus{-1.5} & \tableplus{+2.5} & \cellcolor[HTML]{EFEFEF}\tableplus{+0.9} \\
ER &  & \tableminus{-1.6} & \tableminus{-2.7} & \tableplus{+0.2} & \cellcolor[HTML]{EFEFEF}\tableminus{-1.4} & \tableplus{+19.0} & \tableplus{+13.0} & \tableplus{+20.1} & \cellcolor[HTML]{EFEFEF}\tableplus{+17.4} & \tableplus{+0.3} & \tableminus{-3.5} & \tableplus{+0.9} & \cellcolor[HTML]{EFEFEF}\tableminus{-0.8} \\
LwF &  & \tableminus{-10.7} & \tableminus{\tableminus{-11.9}} & \tableminus{-7.9} & \cellcolor[HTML]{EFEFEF}\tableminus{-10.2} & \tableplus{+12.2} & \tableplus{+2.7} & \tableplus{+10.2} & \cellcolor[HTML]{EFEFEF}\tableplus{+8.3} & \tableminus{-0.8} & \tableminus{-5.4} & \tableplus{+0.8} & \cellcolor[HTML]{EFEFEF}\tableminus{-1.8} \\
EWC &  & \tableminus{\tableminus{-25.9}} & \tableminus{-19.0} & \tableminus{\tableminus{-13.3}} & \cellcolor[HTML]{EFEFEF}\tableminus{\tableminus{-19.4}} & \tableplus{+19.3} & \tableplus{+11.2} & \tableplus{+18.4} & \cellcolor[HTML]{EFEFEF}\tableplus{+16.3} & \tableminus{-9.8} & \tableminus{-8.4} & \tableminus{-4.2} & \cellcolor[HTML]{EFEFEF}\tableminus{-7.5} \\
ECLIPSE &  & \tableminus{-6.0} & \tableminus{-6.2} & \tableminus{-3.9} & \cellcolor[HTML]{EFEFEF}\tableminus{-5.3} & \tableplus{+2.2} & \tableplus{+0.2} & \tableplus{+4.3} & \cellcolor[HTML]{EFEFEF}\tableplus{+2.2} & \tableplus{+0.9} & \tableminus{-3.6} & \tableplus{+2.0} & \cellcolor[HTML]{EFEFEF}\tableminus{-0.3} \\
\textbf{ConOVS (ours)} & \multirow{-6}{*}{\begin{tabular}[c]{@{}c@{}}Cityscapes\end{tabular}} & \textbf{\tableplus{+0.3}} & \textbf{\tableplus{+0.5}} & \textbf{\tableplus{+0.1}} & \cellcolor[HTML]{EFEFEF}\textbf{\tableplus{+0.3}} & \textbf{\tableplus{+20.2}} & \textbf{\tableplus{+13.9}} & \textbf{\tableplus{+21.3}} & \cellcolor[HTML]{EFEFEF}\textbf{\tableplus{+18.5}} & \textbf{\tableplus{+2.5}} & \textbf{\tableminus{-1.2}} & \textbf{\tableplus{+2.5}} & \cellcolor[HTML]{EFEFEF}\textbf{\tableplus{+1.3}} \\ \midrule
X-Decoder & - & 56.7 & 46.9 & 67.4 & \cellcolor[HTML]{EFEFEF}57.0 & 36.3 & 25.4 & 52.9 & \cellcolor[HTML]{EFEFEF}38.2 & 16.7 & 11.7 & 24.9 & \cellcolor[HTML]{EFEFEF}17.8 \\ \midrule
Fine-tuning &  & \tableminus{-50.4} & \tableminus{\tableminus{-32.2}} & \tableminus{\tableminus{-53.7}} & \cellcolor[HTML]{EFEFEF}\tableminus{\tableminus{-45.5}} & \tableplus{+26.6} & \tableplus{+11.7} & \tableplus{+26.7} & \cellcolor[HTML]{EFEFEF}\tableplus{+21.7} & \tableminus{\tableminus{-12.9}} & \tableminus{-8.1} & \tableminus{\tableminus{-19.7}} & \cellcolor[HTML]{EFEFEF}\tableminus{\tableminus{-13.5}} \\
\textbf{ConOVS (ours)} & \multirow{-2}{*}{\begin{tabular}[c]{@{}c@{}}Cityscapes\end{tabular}} & \textbf{\tableminus{-0.4}} & \textbf{\tableminus{-0.4}} & \textbf{\tableminus{-0.3}} & \cellcolor[HTML]{EFEFEF}\textbf{\tableminus{-0.3}} & \textbf{\tableplus{+26.6}} & \textbf{\tableplus{+11.6}} & \textbf{\tableplus{+26.7}} & \cellcolor[HTML]{EFEFEF}\textbf{\tableplus{+21.7}} & \textbf{\tableplus{+0.1}} & \textbf{\tableplus{+0.5}} & \textbf{\tableminus{-0.3}} & \cellcolor[HTML]{EFEFEF}\textbf{\tableplus{+0.1}} \\
\end{tabular}
\label{tab:cityscapes_all}
\end{table}

\begin{table}[h]
\scriptsize
\setlength{\tabcolsep}{0.7em}
\renewcommand{\arraystretch}{1.2}
\caption{\small Performance comparison among the baseline, fine-tuning, retraining, existing continual learning methods, and ConOVS (ours). The incremental dataset is ADE20K. We use \pq, \map, and \miou as evaluation metrics.}
\begin{tabular}{@{}lccccc|cccc|cccc@{}}
 &  & \multicolumn{4}{c|}{\textbf{COCO (pre-training)}} & \multicolumn{4}{c|}{\textbf{ADE20k (incremental)}} & \multicolumn{4}{c}{\textbf{Cityscapes (zero-shot)}} \\
\multirow{-2}{*}{\textbf{Method}} & \multirow{-2}{*}{\textbf{\begin{tabular}[c]{@{}c@{}}Incremental \\ Dataset\end{tabular}}} & \textbf{\pq} & \textbf{\map} & \textbf{\miou} & \cellcolor[HTML]{EFEFEF}\textbf{\texttt{Avg}} & \textbf{\pq} & \textbf{\map} & \textbf{\miou} & \cellcolor[HTML]{EFEFEF}\textbf{\texttt{Avg}} & \textbf{\pq} & \textbf{\map} & \textbf{\miou} & \cellcolor[HTML]{EFEFEF}\textbf{\texttt{Avg}} \\ \Xhline{3\arrayrulewidth}
\addlinespace[3pt]
fc-clip & - & 50.1 & 41.1 & 52.0 & \cellcolor[HTML]{EFEFEF}47.7 & 23.5 & 17.1 & 30.4 & \cellcolor[HTML]{EFEFEF}23.7 & 44.0 & 26.8 & 56.2 & \cellcolor[HTML]{EFEFEF}42.4 \\
\midrule
Fine-tuning &  & \tableminus{-7.7} & \tableminus{-6.2} & \tableminus{-2.7} & \cellcolor[HTML]{EFEFEF}\tableminus{-5.5} & \tableplus{+24.1} & \tableplus{+19.0} & \tableplus{+22.0} & \cellcolor[HTML]{EFEFEF}\tableplus{+21.7} & \tableminus{-3.0} & \tableminus{-2.8} & \tableplus{+2.9} & \cellcolor[HTML]{EFEFEF}\tableminus{-1.0} \\
Retraining &  & \tableplus{+1.4} & \tableplus{+2.2} & \tableplus{+2.9} & \cellcolor[HTML]{EFEFEF}\tableplus{+2.2} & \tableplus{+16.5} & \tableplus{+13.3} & \tableplus{+15.0} & \cellcolor[HTML]{EFEFEF}\tableplus{+14.9} & \tableminus{-1.2} & \tableminus{-0.7} & \tableplus{+1.3} & \cellcolor[HTML]{EFEFEF}\tableminus{-0.2} \\
ER &  & \tableplus{+0.4} & \tableminus{-0.3} & \tableplus{+2.9} & \cellcolor[HTML]{EFEFEF}\tableplus{+1.0} & \tableplus{+21.5} & \tableplus{+16.3} & \tableplus{+19.5} & \cellcolor[HTML]{EFEFEF}\tableplus{+19.1} & \tableminus{-3.5} & \tableminus{-2.8} & \tableminus{-1.0} & \cellcolor[HTML]{EFEFEF}\tableminus{-2.4} \\
LwF &  & \tableminus{-3.8} & \tableminus{-7.1} & \tableminus{-2.4} & \cellcolor[HTML]{EFEFEF}\tableminus{-4.4} & \tableplus{+13.7} & \tableplus{+8.4} & \tableplus{+11.3} & \cellcolor[HTML]{EFEFEF}\tableplus{+11.1} & \tableminus{-1.0} & \tableminus{-6.2} & \tableminus{-3.0} & \cellcolor[HTML]{EFEFEF}\tableminus{-3.4} \\
EWC &  & \tableminus{\tableminus{-11.1}} & \tableminus{-9.3} & \tableminus{-6.0} & \cellcolor[HTML]{EFEFEF}\tableminus{-8.8} & \tableplus{+20.7} & \tableplus{+16.2} & \tableplus{+18.0} & \cellcolor[HTML]{EFEFEF}\tableplus{+18.3} & \tableminus{-2.6} & \tableminus{-3.2} & \tableplus{+0.3} & \cellcolor[HTML]{EFEFEF}\tableminus{-1.8} \\
ECLIPSE &  & \tableminus{-0.5} & \tableminus{-1.2} & \tableplus{+0.6} & \cellcolor[HTML]{EFEFEF}\tableminus{-0.3} & \tableplus{+0.2} & \tableminus{-0.3} & \tableplus{+3.0} & \cellcolor[HTML]{EFEFEF}\tableplus{+1.0} & \tableminus{-5.9} & \tableminus{-4.0} & \tableminus{-2.2} & \cellcolor[HTML]{EFEFEF}\tableminus{-4.0} \\
\textbf{ConOVS (ours)} & \multirow{-6}{*}{\begin{tabular}[c]{@{}c@{}}ADE20K\end{tabular}} & \textbf{\tableplus{+1.7}} & \textbf{\tableplus{+1.4}} & \textbf{\tableplus{+3.2}} & \cellcolor[HTML]{EFEFEF}\textbf{\tableplus{+2.1}} & \textbf{\tableplus{+23.8}} & \textbf{\tableplus{+18.6}} & \textbf{\tableplus{+21.1}} & \cellcolor[HTML]{EFEFEF}\textbf{\tableplus{+21.2}} & \textbf{\tableplus{+0.8}} & \textbf{\tableplus{+0.8}} & \textbf{\tableplus{+1.8}} & \cellcolor[HTML]{EFEFEF}\textbf{\tableplus{+1.1}} \\ \midrule
X-Decoder & - & 56.7 & 46.9 & 67.4 & \cellcolor[HTML]{EFEFEF}57.0 & 16.7 & 11.7 & 24.9 & \cellcolor[HTML]{EFEFEF}17.8 & 36.3 & 25.4 & 52.9 & \cellcolor[HTML]{EFEFEF}38.2 \\ \midrule
Fine-tuning &  & \tableminus{\tableminus{-37.3}} & \tableminus{\tableminus{-33.6}} & \tableminus{\tableminus{-42.4}} & \cellcolor[HTML]{EFEFEF}\tableminus{-37.8} & \tableplus{+28.2} & \tableplus{+18.6} & \tableplus{+27.2} & \cellcolor[HTML]{EFEFEF}\tableplus{+24.6} & \tableminus{-3.7} & \tableminus{-9.4} & \tableminus{-0.8} & \cellcolor[HTML]{EFEFEF}\tableminus{-4.6} \\
\textbf{ConOVS (ours)} & \multirow{-2}{*}{\begin{tabular}[c]{@{}c@{}}ADE20K \end{tabular}} & \textbf{\tableminus{-1.5}} & \textbf{\tableminus{-1.7}} & \textbf{\tableminus{-1.1}} & \cellcolor[HTML]{EFEFEF}\textbf{\tableminus{-1.4}} & \textbf{\tableplus{+29.2}} & \textbf{\tableplus{+19.0}} & \textbf{\tableplus{+27.5}} & \cellcolor[HTML]{EFEFEF}\textbf{\tableplus{+25.2}} & \textbf{\tableplus{+1.4}} & \textbf{\tableminus{-6.4}} & \textbf{\tableplus{+3.5}} & \cellcolor[HTML]{EFEFEF}\textbf{\tableminus{-0.5}} \\

\end{tabular}
\label{tab:ade20k_all}
\end{table}
\begin{table}[h]
\scriptsize
\setlength{\tabcolsep}{1.0em}
\renewcommand{\arraystretch}{1.2}
\caption{
\small Performance comparison among the baseline, fine-tuning, retraining, existing continual learning methods, and ConOVS (ours). The underlined values indicate the best score for each dataset. We use \pq, \map, and \miou as evaluation metrics.}
\begin{tabular}{@{}lcccc|ccc|ccc@{}}
 &  & \multicolumn{3}{c|}{\begin{tabular}[c]{@{}c@{}}\textbf{COCO (pre-training)}\end{tabular}} & \multicolumn{3}{c|}{\begin{tabular}[c]{@{}c@{}}\textbf{ADE20k (incremental)}\end{tabular}} & \multicolumn{3}{c}{\begin{tabular}[c]{@{}c@{}}\textbf{Cityscapes (incremental)}\end{tabular}} \\
\multirow{-2}{*}{\textbf{Method}} & \multirow{-2}{*}{\textbf{\begin{tabular}[c]{@{}c@{}}Learning Sequence\end{tabular}}} & \textbf{\pq} & \textbf{\map} & \textbf{\miou}  & \textbf{\pq} & \textbf{\map} & \textbf{\miou}& \textbf{\pq} & \textbf{\map} & \textbf{\miou} \\ \midrule \midrule
fc-clip & - & 50.1 & 41.1 & 52.0 & 23.5 & 17.1 & 30.4 & 44.0 & 26.8 & 56.2 \\
Fine-tuning & ADE20k → Cityscapes & 20.8 & 19.5 & 40.0 & 15.4 & 14.2 & 34.9 & {\ul 65.2} & {\ul 42.3} & {\ul 77.6} \\
Fine-tuning & Cityscapes → ADE20k & 39.3 & 32.4 & 48.3 & {\ul 48.3} & {\ul 36.3} & {\ul 52.1} & 46.0 & 26.4 & 61.5 \\
Retraining & COCO, Cityscapes, ADE20k & 48.6 & 40.5 & 50.9 & 35.5 & 25.8 & 38.3 & 60.5 & 36.5 & 74.3 \\
ConOVS (ours) & Cityscapes, ADE20k & {\ul \textbf{51.6}} & {\ul \textbf{42.5}} & {\ul \textbf{55.3}} & \textbf{47.0} & \textbf{35.9} & \textbf{51.4} & \textbf{64.3} & \textbf{40.7} & {\ul \textbf{77.6}}
\end{tabular}
\label{tab:multiple_all}
\end{table}

\end{document}